\documentclass[sn-mathphys]{sn-jnl}

\newcommand{\Real}{\mathbb{R}}
\newcommand{\BB}{\boldsymbol}

\definecolor{fhcolor}{rgb}{0.523, 0.235, 0.625}

\jyear{2022}%

\theoremstyle{thmstyleone}%
\newtheorem{thm}{Theorem}
\newtheorem{proposition}[thm]{Proposition}%
\newtheorem{corollary}[thm]{Corollary}

\theoremstyle{thmstyletwo}%
\newtheorem{remark}{Remark}%
\newtheorem*{prf}{Proof}

\theoremstyle{thmstylethree}%
\newtheorem{definition}{Definition}%

\begin{document}

\title[Random Fourier Features for Asymmetric kernels]{Random Fourier Features for Asymmetric kernels}


\author[1]{\fnm{Mingzhen} \sur{He}}\email{mingzhen\_he@sjtu.edu.cn}

\author[1]{\fnm{Fan} \sur{He}}\email{hf-inspire@sjtu.edu.cn}

\author[2]{\fnm{Fanghui} \sur{Liu}}\email{fanghui.liu@epfl.ch}
\author*[1]{\fnm{Xiaolin} \sur{Huang}}\email{xiaolinhuang@sjtu.edu.cn}

\affil[1]{\orgdiv{Institute of Image Processing and Pattern Recognition}, \orgname{Shanghai Jiao Tong University}, \orgaddress{\street{800 Dongchuan RD. Minhang District}, \city{Shanghai}, \postcode{200240}, \country{China}}}

\affil[2]{\orgdiv{Laboratory for Information and Inference Systems}, \orgname{École Polytechnique Fédérale de Lausanne}, \orgaddress{\street{EPFL SV-BMC AA-B118}, \city{Lausanne}, \postcode{1015}, \country{Switzerland}}}


\abstract{The random Fourier features (RFFs) method is a powerful and popular technique in kernel approximation for scalability of kernel methods. The theoretical foundation of RFFs is based on the Bochner theorem \citep{bochner2005harmonic} that relates symmetric, positive definite (PD) functions to probability measures.
This condition naturally excludes asymmetric functions with a wide range applications in practice, e.g., directed graphs, conditional probability, and asymmetric kernels. 
Nevertheless, understanding asymmetric functions (kernels) and its scalability via RFFs is unclear both theoretically and empirically. 
In this paper, we introduce a complex measure with the real and imaginary parts corresponding to four finite positive measures, which expands the application scope of the Bochner theorem. 
By doing so, this framework allows for handling classical symmetric, PD kernels via one positive measure; symmetric, non-positive definite kernels via signed measures; and asymmetric kernels via complex measures, thereby unifying them into a general framework by RFFs, named AsK-RFFs.
Such approximation scheme via complex measures enjoys theoretical guarantees in the perspective of the uniform convergence.
In algorithmic implementation, to speed up the kernel approximation process, which is expensive due to the calculation of total mass, we employ a subset-based fast estimation method that optimizes total masses on a sub-training set, which enjoys computational efficiency in high dimensions.
Our AsK-RFFs method is empirically validated on several typical large-scale datasets and achieves promising kernel approximation performance, which demonstrate the effectiveness of AsK-RFFs.}




\maketitle

\section{Introduction}
Kernel methods \citep{scholkopf2002learning, suykens2002least, vapnik2013nature} play a significant role in statistical machine learning in almost all the topics, e.g., classification \citep{soman2009machine}, regression \citep{drucker1997support}, clustering \citep{girolami2002mercer}, and dimension reduction \citep{scholkopf1997kernel}. 
One significant limitation is the lack of scalability due to the fact that kernel methods are working on the kernel matrices which have the size $N \times N$ with $N$ being the number of samples. 
Kernel approximation \citep{Williams01usingthe,drineas2005nystrom,liu2021random} is able to address this issue by approximating the original kernel matrix or kernel function.
One typical technique tool is
\emph{Random Fourier features} (RFFs) that introduces low dimensional randomized feature mappings for approximating the original kernel function \citep{rahimi2007random,rahimi2008weighted}. The theoretical foundation of RFFs is established by the Bochner's theorem in harmonic analysis \citep{bochner2005harmonic,rahimi2007random}. The Bochner's theorem claims that a continuous, symmetric, shift-invariant and positive definite (PD) function can be approximated by a finite non-negative measure.
Besides, RFFs can be regarded as a two-layer neural network with only training the second layer, which naturally bridges kernel methods and neural networks in deep learning theory \citep{mehrkanoon2018deep,belkin2019reconciling,malach2020proving}; refer to one recent survey \citep{liu2021random}. 

The target of RFFs is to establish an explicit feature mapping via the sampled random features and then approximate the original kernel function via inner product. 
There is a strict pre-condition on RFFs that the sampling measure is well-defined only when the kernel is stationary, PD, and symmetric. 
Extension of RFFs to 
other kernels, e.g., polynomial kernels \citep{pennington2015spherical}, non-positive definite kernel, e.g., the neural tangent kernel of two-layer ReLU networks with unit spherical constraints \citep{liu2021fast} is important and has been discussed. 
However, study on RFFs for asymmetric kernels is missing in the machine learning community.

In practice, asymmetric similarities/dissimilarities are commonly used.
Here we take several typical examples to illustrate this.
1) In directed graphs, the adjacency matrices that record (weighted) 
connections between two samples can efficiently represent asymmetric similarities. 
The relationship between a central and a local node can be asymmetric. For example, in federated learning, the central update is aggregated by updates of all local nodes.
In this case, neural tangent kernel-based federated learning is asymmetric \citep{huang2021fl}. 
2) the conditional probability is another popular scene we may meet asymmetric similarities due to ${\rm P}(A\vert B)\neq {\rm P}(B\vert A)$ in general, where $A$ and $B$ are two events in an event space. Some applications for conditional probabilities could be found in the works of \cite{hinton2002stochastic,tang2015pte,tang2015line}.
3) Asymmetric kernels, as a tool more flexible than PD kernels, are promising to connect to deep models. As an encouraging attempt, \cite{wright2021transformers} successfully characterize the dot-product attention by asymmetric kernels and provides a theoretical understanding of transformer models \citep{vaswani2017attention}. 

In theory, the function space associated with asymmetric kernels is often built on reproducing kernel Banach spaces (RKBS) \citep{zhang2009reproducing,song2013reproducing,fasshauer2015solving,bartolucci2021understanding,lin2022reproducing} instead of the classical reproducing kernel Hilbert spaces (RKHS).  
For the primal feature analysis,  \cite{suykens2016svd,suykens2017deep} introduce a variational principle for understanding the singular value decomposition from the perspective of asymmetric kernels associated with different feature mappings. 
These theories are helpful for understanding asymmetric kernels but are not suitable to handle the 
approximation problem.

Extension of standard RFFs to asymmetric kernels is unanswered 
and non-trivial both theoretically and algorithmically.
To be specific, theoretically, the original probability measure for PD kernels \citep{rahimi2007random} and signed measure for indefinite but symmetric kernels \citep{HUANG2017162,liu2021fast} cannot be used as
the sampling measure for asymmetric kernels, which is the main obstacle of RFFs for asymmetric kernels.
Algorithmically, the point evaluation at the origin 
is not sufficient to represent the corresponding total masses of measures, which differs from that in the Bochner's theorem. In this scene, calculation of total masses often involves a $d$ dimensional numerical integral, which is time-consuming in high dimensions.

In this paper, we propose a unified kernel approximation framework for asymmetric kernels by RFFs, called \emph{AsK-RFFs}, which fills the gap of asymmetric kernels from theory to practice. 
Ask-RFFs design a complex measure on the sample space for a real, absolutely integrable, and shift-invariant kernel, without any requirement on symmetry. In theory, we provide the uniform convergence bound of the proposed AsK-RFFs, which is not only for asymmetric kernels but is also applicable to PD and indefinite kernels. In the algorithm, we partition a complex measure into four finite positive measures and then approximate asymmetric kernels by using the Monte Carlo sampling method to sample a series of random Fourier features. 
One crucial and skillful point is to efficiently calculate total masses via a subset-based strategy.
By doing so, Ask-RFFs enjoys nice theoretical guarantees and promising approximation performance in several large-scale benchmark datasets.


In the rest of this paper, Section \ref{sec:prelimiary} briefly introduces preliminaries about RFF 
and measures. Then, we give the formulation of AsK-RFFs with its uniform convergence bound in Section \ref{sec:method}. We give three typical asymmetric kernels and investigate an estimation algorithm for total masses in Section \ref{sec:Algo_examples}. Section \ref{sec:experiment} shows the experimental results of the proposed AsK-RFFs on several large datasets. Finally, a brief conclusion is drawn in Section \ref{sec:conclusion}.

\section{Preliminaries}\label{sec:prelimiary}
In this section, we briefly sketch several basic properties of random Fourier features and complex measures which are needful in the approximation for asymmetric kernels. Let $\mathcal{X} = \{\BB{x_k}\}_{k=1}^N \subset\Real^d$ be the sample set.
Denote $\mathsf{i}=\sqrt{-1}$.

\begin{thm}[Bochner's theorem \citep{bochner2005harmonic}]\label{Bochner_thm} Let $k:\Real^d\times\Real^d\rightarrow\Real$ be a continuous, symmetric, and shift-invariant function, i.e., $k(\BB{x},\BB{y})=k(\BB{x}-\BB{y})$ and $k(\BB{0})=1$. Then $k$ is a positive definite function if and only if it is the Fourier transform of a finite non-negative Borel measure $\mu$.
\begin{equation*}
    k(\BB{x}-\BB{y}) = \int_{\Real^d} e^{\mathsf{i}\BB{\omega}^\top(\BB{x}-\BB{y})} \mu(\mathrm{d}\BB{\omega}) = \mathbb{E}_{\BB{\omega}\sim\mu}\left[e^{\mathsf{i}\BB{\omega}^\top(\BB{x}-\BB{y})}\right]\,.
\end{equation*}
\end{thm}

A positive definite kernel can be constructed by a finite non-negative Borel measure. According to Theorem \ref{Bochner_thm}, one can approximate a positive definite (symmetric) kernel $k(\BB{x}-\BB{y})=\mathbb{E}_{\BB{\omega}\sim\mu}\left[e^{i\BB{\omega}^\top(\BB{x}-\BB{y})}\right]$ by 
the Monte Carlo sampling method to sample a series of random Fourier features $\{\BB{\omega_k}\}_{k=1}^M$ from the normalized  measure $\mu$ which can be regarded as a probability measure by taking $k(\bm 0) = 1$ \citep{rahimi2007random}.
However, for asymmetric kernels, their Fourier transform can not be associated with a probability measure, which does not allows for designing the random features sampling strategy in our problem.
To address this, we need study the conjugate property and complex measure that are required for analysis of non-positive definite kernels.

\begin{thm}[Conjugate property \citep{Fourier_analysis_2008}]\label{conjugate_property}
Given a real absolutely integrable function $k(\BB{\Delta})$ and its Fourier transform $\mu(\BB{\omega})$ where $\BB{\Delta},\BB{\omega}\in\Real^d$, then one has the reality condition $\mu(-\BB{\omega}) = \overline{\mu(\BB{\omega})}$.
\end{thm}
Denote $\mu:=\mu_R+\mathsf{i}\mu_I$, then it is obvious that $\mu_R$ is even and $\mu_I$ is odd with respect to $\BB{\omega}$, i.e., $\mu_R(\BB{\omega}) = \mu_R(-\BB{\omega})$ and $\mu_I(\BB{\omega}) = -\mu_I(-\BB{\omega})$.
In the following, we introduce complex measures which will help to build the asymmetric kernel approximation. Actually, a finite positive measure and a finite signed measure have been used in approaching PD
kernels \citep{rahimi2007random} and indefinite kernels \citep{liu2021fast}, respectively.


\begin{definition}[Complex measures \citep{cohn2013measure}]\label{complex_measure}
Let $\Omega$ be a sample set, $\mathcal{M}$ be a $\sigma$-algebra of subsets on $\Omega$ and $(\Omega, \mathcal{M})$ be a measurable space. A complex measure on $(\Omega, \mathcal{M})$ is a complex-valued function $\mu:\mathcal{M}\rightarrow\mathbb{C}$ that satisfies $\mu(\varnothing)=0$ and countable additivity.
\end{definition}

A complex measure $\mu$ on a measurable space $(\Omega, \mathcal{M})$ can be formulated as $\mu=\mu_R+\mathsf{i}\mu_I$, where $\mu_R$ and $\mu_I$ are two finite signed measures on $(\Omega, \mathcal{M})$. 


\begin{thm}[Jordan decomposition \citep{athreya2006measure}]\label{jordan}
Every signed measure is the difference of two positive measures $\zeta=\zeta^+ - \zeta^-$, at least one of which is finite.
\end{thm}

The Jordan decomposition theory implies that a complex measure $\mu$ consists of four finite positive measures $\mu_R^+$, $\mu_R^-$, $\mu_I^+$ and $\mu_I^-$ on a measurable space $(\Omega, \mathcal{M})$ as the following,
\begin{equation}\label{complex_decomposition}
    \mu = \mu_R^+ - \mu_R^- +\mathsf{i}\mu_I^+ - \mathsf{i}\mu_I^-\,.
\end{equation}
Notice that the decomposition of a complex measure is not unique.

\section{Methods}\label{sec:method}
In this section, we propose a novel framework for 
approximating asymmetric kernels by random Fourier features (AsK-RFFs). We will first introduce the random features 
and then establish a uniform convergence on the approximation ability. 
The algorithms including the sampling method and the total mass estimation are given 
in the next section. 

Given a real and absolutely integrable kernel function $k:\Real^d\times\Real^d\rightarrow\Real$, $k(\BB{x},\BB{y})=k(\BB{x}-\BB{y})=k(\BB{\Delta})$ (maybe not symmetric), $\BB{x},\BB{y}\in\mathcal{X}\subset\Real^d$ and its Fourier transform as a complex measure $\mu(\BB{\omega})=\mu_R(\BB{\omega})+\mathsf{i}\mu_I(\BB{\omega})$ where $\BB{\omega}\in\Real^d$, we can define four positive measures as the following,
\begin{equation}\label{positive_measures}
    \left\{
    \begin{aligned}{}
        &\mu_R^+= \max\{\mu_R ,0\},~  \mu_R^- = \max\{-\mu_R,0\}\\
        &\mu_I^+= \max\{\mu_I ,0\},~\mu_I^- = \max\{-\mu_I,0\}\,.\\
    \end{aligned}
    \right.
\end{equation}
According to Theorem \ref{conjugate_property}, the imaginary part $\mu_I$ is odd, and then the following equation holds,
\begin{equation}\label{p1_imaginarypart}
    \mu_I^+(-\BB{\omega}) = \max\{\mu_I(-\BB{\omega}),0\}=\max\{-\mu_I(\BB{\omega}),0\}=\mu_I^-(\BB{\omega})\,.
\end{equation}

\begin{definition}(Total masses of positive measures)\label{TV_pm}
Suppose $\zeta$ is a finite positive measure on a measurable space $(\Omega, \mathcal{M})$, the total mass of $\zeta$, denoted $\|\zeta\|$, is defined by $\|\zeta\| =\int_{\BB{\omega}\in\Omega} \zeta(\mathrm{d}\BB{\omega}) = \zeta(\Omega) < \infty$.
\end{definition}
\begin{remark}
Specifically, a (normalized) measure $\Tilde{\zeta}=\zeta/\|\zeta\|$ is a probability measure i.e., $\Tilde{\zeta}(\Omega) = 1$. Then a mathematical triple $(\Omega, \mathcal{M},\Tilde{\zeta})$ is the corresponding probability space. It helps us to ``sample'' from a finite positive measure $\zeta$, $\int_\Omega fd\zeta=\int_\Omega f\|\zeta\|d\Tilde{\zeta}=\mathbb{E}_{\Tilde{\zeta}} [f\|\zeta\|]$.
\end{remark}

According to the Jordan decomposition, a complex measure is made up of four finite positive measures, as shown in (\ref{complex_decomposition}). In this paper, the total mass of a complex measure is defined as follows,

\begin{definition}(Total masses of complex measures)\label{TV_cm}
Suppose a complex measure $\mu$ on a measurable space $(\Omega, \mathcal{M})$ can be written in the form of (\ref{complex_decomposition}) i.e., $\mu = \mu_R^+ - \mu_R^- +\mathsf{i}\mu_I^+ - \mathsf{i}\mu_I^-$. 
The total mass of $\mu$, denoted $\|\mu\|$, is defined by
\begin{equation}
    \|\mu\| = \|\mu_R^+\|+\|\mu_R^-\|+\|\mu_I^+\|+\|\mu_I^-\|\,.
\end{equation}
\end{definition}

Then, we begin to introduce the approximation process of the proposed AsK-RFFs. Firstly, a real, absolutely integrable, shift-invariant kernel function $k$ can be represented by the following proposition.
\begin{proposition}[Representation of kernels by complex measures]\label{kernel_represent}
A real, absolutely integrable, shift-invariant kernel function $k$ can be represented by its Fourier transform $\mu$ with $\|\mu\|<\infty$ which can be regarded as a complex measure on a measurable space $(\Omega, \mathcal{M})$, where $\mu_{R}^+,\mu_{R}^-,\mu_{I}^+,\mu_{I}^-$ are defined by \eqref{positive_measures},
\begin{equation*}
    \begin{aligned}
    k(\BB{\Delta})&=\int_{\Real^d} \cos(\BB{\omega}^\top\BB{\Delta})\mu_{R}^+(\mathrm{d}\BB{\omega})-\int_{\Real^d} \cos(\BB{\omega}^\top\BB{\Delta})\mu_{R}^-(\mathrm{d}\BB{\omega})\\
    &-\int_{\Real^d}\sin(\BB{\omega}^\top\BB{\Delta})\mu_{I}^+(\mathrm{d}\BB{\omega})+\int_{\Real^d}\sin(\BB{\omega}^\top\BB{\Delta})\mu_{I}^-(\mathrm{d}\BB{\omega})\,.\\
    \end{aligned}
\end{equation*}
\end{proposition}
\begin{prf}
An inverse Fourier transform between a kernel function 
and the complex measure 
can be formulated as follows,
\begin{equation}\label{k_delta}
    \begin{aligned}
        k(\BB{\Delta}) &=\int_{\Real^d} e^{\mathsf{i}\BB{\omega}^\top\BB{\Delta}}\mu(\mathrm{d}\BB{\omega}) 
        =\int_{\Real^d}e^{\mathsf{i}\BB{\omega}^\top\BB{\Delta}}\mu_R(\mathrm{d}\BB{\omega}) + \mathsf{i}\int_{\Real^d}e^{\mathsf{i}\BB{\omega}^\top\BB{\Delta}}\mu_I(\mathrm{d}\BB{\omega})\\
        &=\int_{\Real^d} \left[\cos(\BB{\omega}^\top\BB{\Delta}) + \mathsf{i}\sin(\BB{\omega}^\top\BB{\Delta})\right]\mu_R(\mathrm{d}\BB{\omega}) +  \int_{\Real^d}\left[\mathsf{i}\cos(\BB{\omega}^\top\BB{\Delta})-\sin(\BB{\omega}^\top\BB{\Delta})\right]\mu_I(\mathrm{d}\BB{\omega})\\
        &=\int_{\Real^d} \cos(\BB{\omega}^\top\BB{\Delta})\mu_R(\mathrm{d}\BB{\omega})-\int_{\Real^d}\sin(\BB{\omega}^\top\BB{\Delta})\mu_I(\mathrm{d}\BB{\omega})+ \mathsf{i} \int_{\Real^d}\sin(\BB{\omega}^\top\BB{\Delta})\mu_R(\mathrm{d}\BB{\omega}) \\
        &\quad+ \mathsf{i}\int_{\Real^d}\cos(\BB{\omega}^\top\BB{\Delta})\mu_I(\mathrm{d}\BB{\omega})\\
        &= \int_{\Real^d} \cos(\BB{\omega}^\top\BB{\Delta})\mu_R(\mathrm{d}\BB{\omega})-\int_{\Real^d}\sin(\BB{\omega}^\top\BB{\Delta})\mu_I(\mathrm{d}\BB{\omega})\\
        &=\int_{\Real^d} \cos(\BB{\omega}^\top\BB{\Delta})\mu_{R}^+(\mathrm{d}\BB{\omega})-\int_{\Real^d} \cos(\BB{\omega}^\top\BB{\Delta})\mu_{R}^-(\mathrm{d}\BB{\omega})-\int_{\Real^d}\sin(\BB{\omega}^\top\BB{\Delta})\mu_{I}^+(\mathrm{d}\BB{\omega})\\
        &\quad+\int_{\Real^d}\sin(\BB{\omega}^\top\BB{\Delta})\mu_{I}^-(\mathrm{d}\BB{\omega})\,.\\
    \end{aligned}
\end{equation}
The fourth equation holds because $k$ is a real kernel thus the imaginary part of the representation can be ignored.
\end{prf}

According to (\ref{k_delta}), 
there are two constraints on $\mu$ for being the complex measure of a kernel.
\begin{corollary}[Constraints of complex measures]\label{constrain_complex_measures}
Suppose $\mu = \mu_R^+ - \mu_R^- +\mathsf{i}\mu_I^+ - \mathsf{i}\mu_I^-$ is a Fourier transform of a real kernel function $k$ and the total mass $\|\mu\|$ is finite, then 
\begin{equation}\label{constrain_complex_measures_algo}
    \left\{
    \begin{aligned}{}
        &\|\mu_R^+\| - \|\mu_R^-\| = k(\BB{0})\\
        &\|\mu_I^+\| - \|\mu_I^-\| = 0.\,\\
    \end{aligned}
    \right.
\end{equation}
\end{corollary}
\begin{prf}
Substituting $\BB{\Delta}=\BB{0}$ into (\ref{k_delta}), one can obtain that
\begin{equation*}
    \begin{aligned}
        k(\BB{0}) &=\int_{\Real^d} e^{\mathsf{i}\BB{\omega}^\top\BB{0}}\mu(\mathrm{d}\BB{\omega})=\int_{\Real^d} \mu_R(\mathrm{d}\BB{\omega})+ \mathsf{i}\int_{\Real^d}\mu_I(\mathrm{d}\BB{\omega})\\
        &=\int_{\Real^d} \mu_R^+(\mathrm{d}\BB{\omega})-\int_{\Real^d} \mu_R^-(\mathrm{d}\BB{\omega})+ \mathsf{i}\int_{\Real^d}\mu_I^+(\mathrm{d}\BB{\omega})-\mathsf{i}\int_{\Real^d}\mu_I^-(\mathrm{d}\BB{\omega})\\
        &=\|\mu_R^+\| - \|\mu_R^-\| + \mathsf{i}\left(\|\mu_I^+\| - \|\mu_I^-\|\right).\,\\
    \end{aligned}
\end{equation*}
$k$ is a real kernel thus $k(\bm 0)$ is of course real. 
In other words, the imaginary part is zero, and thus (\ref{constrain_complex_measures_algo}) should hold.
\end{prf}
\begin{remark}
Corollary \ref{constrain_complex_measures} provides two rules for a complex measure $\mu$ to be 
a Fourier transform of a real kernel function $k$. According to Definition \ref{TV_cm}, the total mass of the measure $\mu$ can also be defined by $\|\mu\| = \|\mu_R^+\|+\|\mu_R^-\|+2\|\mu_I^+\|$.
\end{remark}

According to (\ref{p1_imaginarypart}), i.e., $\mu_I^+(-\BB{\omega}) = \mu_I^-(\BB{\omega})$, we know that $(\mu_I^++\mu_I^-)(\BB{\omega})$ is an even function with respect to $\BB{\omega}$, i.e., $(\mu_I^++\mu_I^-)(-\BB{\omega})=\mu_I^+(-\BB{\omega})+\mu_I^-(-\BB{\omega})=\mu_I^-(\BB{\omega})+\mu_I^+(\BB{\omega})=(\mu_I^++\mu_I^-)(\BB{\omega})$.
Then, $\sin(\BB{\omega}^\top\BB{\Delta})\cdot\big[\mu_{I}^+(\mathrm{d}\BB{\omega})+\mu_{I}^-(\mathrm{d}\BB{\omega})\big]$ is an odd function with respect to $\BB{\omega}$. Thus we have $\int_{\Real^d}\sin(\BB{\omega}^\top\BB{\Delta})\big[\mu_{I}^+(\mathrm{d}\BB{\omega})+\mu_{I}^-(\mathrm{d}\BB{\omega})\big] = 0$,
which indicates that
\[
\int_{\Real^d}\sin(\BB{\omega}^\top\BB{\Delta})\mu_{I}^+(\mathrm{d}\BB{\omega})= -\int_{\Real^d}\sin(\BB{\omega}^\top\BB{\Delta})\mu_{I}^-(\mathrm{d}\BB{\omega})\,.
\] 
The contributions of the positive imaginary measure $\mu_{I}^+$ and the negative imaginary measure $\mu_{I}^-$ are equivalent. Secondly, we can speed up the approximation process since only one of $\mu_{I}^+$ and $\mu_{I}^-$ need to be sampled.
\begin{equation}\label{kernel_approx}
    \begin{aligned}
        k(\BB{\Delta})
        &=\int_{\Real^d} \cos(\BB{\omega}^\top\BB{\Delta})\mu_{R}^+(\mathrm{d}\BB{\omega})-\int_{\Real^d} \cos(\BB{\omega}^\top\BB{\Delta})\mu_{R}^-(\mathrm{d}\BB{\omega})-2\int_{\Real^d}\sin(\BB{\omega}^\top\BB{\Delta})\mu_{I}^+(\mathrm{d}\BB{\omega})\\
        &=\mathbb{E}_{\BB{\omega}\sim \tilde{\mu}_R^+}\left[\|\mu_{R}^+\|\cos(\BB{\omega}^\top\BB{\Delta})\right]-\mathbb{E}_{\BB{\zeta}\sim \tilde{\mu}_R^-}\left[\|\mu_{R}^-\|\cos(\BB{\zeta}^\top\BB{\Delta})\right]\\
        &\quad-2\mathbb{E}_{\BB{\nu}\sim \tilde{\mu}_I^+}\left[\|\mu_{I}^+\|\sin(\BB{\nu}^\top\BB{\Delta})\right]\\
        &:=k_R^+(\BB{\Delta})-k_R^+(\BB{\Delta})-k_I(\BB{\Delta})\\
        &\approx \|\mu_{R}^+\|\phi(\BB{\omega},\BB{x})^\top\phi(\BB{\omega},\BB{y}) - \|\mu_{R}^-\|\phi(\BB{\zeta},\BB{x})^\top\phi(\BB{\zeta},\BB{y}) - 2\|\mu_{I}^+\|\phi(\BB{\nu},\BB{x})^\top\psi(\BB{\nu},\BB{y})\\
        &:=s(\BB{\Delta})\,, \\
    \end{aligned}
\end{equation}
where $k_R^+$ and $k_R^+$ are two symmetric PD functions, $k_I$ is a skew-symmetric function,  $\tilde{\mu}_R^+,\tilde{\mu}_R^-, \tilde{\mu}_I^+$ are three normalized measures corresponding to $\mu_{R}^+$, $\mu_{R}^-$ and $\mu_{I}^+$, respectively. The feature mapping $\phi(\BB{\omega},\BB{x}),\psi(\BB{\omega},\BB{y})$, $\{\BB{\omega}_k\}_{k=1}^M\sim\Tilde{\mu}$ can be sampled by the Monte Carlo sampling method, where $M$ is the sampling number. Specifically, 
\begin{equation}
    \left\{
    \begin{aligned}{}
         &\phi(\BB{\omega},\BB{x}) = \frac{1}{\sqrt{M}}\left[\cos({\BB{\omega}}_1^\top \BB{x}),\cdots, \cos({\BB{\omega}}_M^\top \BB{x}), \sin({\BB{\omega}}_1^\top \BB{x}),\cdots, \sin({\BB{\omega}}_M^\top \BB{x})\right]^\top \\
         &\psi(\BB{\omega},\BB{y}) = \frac{1}{\sqrt{M}}\left[ -\sin({\BB{\omega}}_1^\top \BB{y}),\cdots, -\sin({\BB{\omega}}_M^\top \BB{y}),\cos({\BB{\omega}}_1^\top \BB{y}),\cdots, \cos({\BB{\omega}}_M^\top \BB{y})\right]^\top\,.\\
    \end{aligned}
    \label{random_features}
    \right.
\end{equation}

As a general framework, Ask-RFFs also covers symmetric kernels, including PD and indefinite ones. In that case, 
the complex measures 
reduce to real signed measures, i.e., $\mu_I = 0$. 
Then the proposed AsK-RFF is equivalent to that in \cite{liu2021fast}, where the kernels to be approximated are symmetric. When the positive definiteness is imposed, the complex measure further degenerates to  a finite real and positive measure. 
In this sense, 
the Bochner’s theorem \citep{bochner2005harmonic} can be regarded as a special case of the AsK-RFFs.

As a kernel approximation method, 
the approximation error with respect to the feature dimension is the most concerned. We give and prove the following uniform convergence for Ask-RFFs. 
\begin{proposition}[Uniform convergence of RFFs for asymmetric kernels]\label{Proposition1}
Let k be a real and continuous kernel function, $k(\BB{x},\BB{y})=k(\BB{\Delta})$ and $\mu=\mu_R^+-\mu_R^- + i(\mu_I^+-\mu_I^-)$ be its Fourier transform with $\|\mu\|=\|\mu_R^+\|+\|\mu_R^-\|+2\|\mu_I^+\|<\infty$, defined on a compact subset $\mathcal{X}\subset\Real^d$ with diameter $l$. Let the approximation function $s(\BB{x},\BB{y})=s(\BB{\Delta})$ be as in (\ref{kernel_approx}) and $\sigma_\mu^2 := \mathbb{E}_{\omega\sim\tilde{\mu}_R^+}\|\omega\|^2+\mathbb{E}_{\zeta\sim\tilde{\mu}_R^-}\|\zeta\|^2 + 2\mathbb{E}_{\nu\sim\tilde{\mu}_I^+}\|\nu\|^2$. Let $\alpha_\mu^2 = \left(\|\mu_R^+\|^2+\|\mu_R^-\|^2+2\|\mu_I^+\|^2\right)\cdot\sigma_\mu^2$ and $\beta_d = \left((\frac{d}{2})^{\frac{-d}{d+2}}+(\frac{d}{2})^{\frac{2}{d+2}}\right)\cdot2^{\frac{6d+2}{d+2}}$.
Then, for any $\epsilon>0$, 
then $\| s - k \|_{\infty}$ is at most $\epsilon$ with probability at least $1-\delta$ as long as
\begin{equation*}
    \begin{aligned}{}
    M \geq \frac{4(d+2)\|\mu\|^2}{\epsilon^2}\left(\log\frac{\beta_d}{\delta}+\frac{2d}{d+2}\log\frac{\alpha_\mu l}{\epsilon}\right)\,.
    \end{aligned}
\end{equation*}
\end{proposition}
The proof sketch 
follows \cite{rahimi2007random, error2015rff} but differs in the considered complex measure. The details
can be found in Appendix \ref{append:proposition1}. The one sentence conclusion is: Pointwise error of Ask-RFF is no more than $\epsilon$ almost surely if the number of random Fourier features $M=\mathcal{O}\left(d\|\mu\|^2\epsilon^{-2}\log\frac{\alpha_\mu l}{\epsilon}\right)$.

\section{Computational Aspects}\label{sec:Algo_examples}
The overall process of Ask-RFFs is summarized in Algorithm \ref{alg:A}, which allows for approximating asymmetric kernels by random Fourier features.
The measures $\mu$, $\mu_R^+$, $\mu_R^-$, and $\mu_I^+$ and their total masses are kernel-dependent, which can be calculated in advance. By this means, the proposed method has the same complexity as RFFs with PD and indefinite kernels \citep{rahimi2007random,liu2021fast}: 
$\mathcal{O}\left(NM^2\right)$ time cost and $\mathcal{O}\left(NM\right)$ memory. 

\begin{algorithm}[htb]
\renewcommand{\algorithmicrequire}{\textbf{Input:}}
\renewcommand{\algorithmicensure}{\textbf{Output:}}
\caption{AsK-RFFs via complex measures.}
\begin{algorithmic}[1]
\Require A kernel function $k(\BB{x},\BB{y})=k(\BB{u})$ with $\BB{u}:=\BB{x}-\BB{y}$ and \#random features $M$.
\Ensure The random Fourier features $\phi,\psi:\Real^d\times\Real^d\rightarrow\Real^{2M}$.
\State Obtain a complex measure $\mu=\mu_R+\mathsf{i}\mu_I$ of the kernel $k$ via the Fourier transform.
\State Let $\mu_R=\mu_R^+-\mu_R^-$ and $\mu_I=\mu_I^+-\mu_I^-$ with four finite positive measures $\mu_R^+$, $\mu_R^-$, $\mu_I^+$ and $\mu_I^-$ 
\State Sample $\{\omega_k\}_{k=1}^M\sim \tilde{\mu}_R^+$, $\{\zeta_k\}_{k=1}^M\sim \tilde{\mu}_R^-$ and $\{\nu_k\}_{k=1}^M\sim \tilde{\mu}_I^+$.
\State Return explicit feature mappings $\phi,\psi$ given in (\ref{random_features}).
\end{algorithmic} \label{alg:A}  
\end{algorithm} 

Beyond the general algorithm, there are some tricky but important details for AsK-RFFs. To better demonstrate the computational details, we first list three practical asymmetric kernels. In the first kernel, the asymmetric part comes from shift and for the last two, the functions are the product of a symmetric part and an asymmetric part. 

\begin{itemize}
    \item \textbf{Shift-Gaussian kernels} 
    are generalized from Gaussian kernels by shifting 
$\BB{r}$ in the time domain. The resulted kernel function and its Fourier transform are given as follows,
\begin{equation}
  \begin{aligned}\label{k1}
        k_1(\BB{\Delta}) & = \exp(-\frac{\|\BB{\Delta}+\BB{r}\|^2}{2\sigma^2})\\
        \mu_1(\BB{\omega}) &= \big(\frac{\sigma}{\sqrt{2\pi}}\big)^d\exp(-\frac{\sigma^2\|\BB{\omega}\|^2}{2})\exp(\mathsf{i}\cdot\BB{r}^\top\BB{\omega})\,,
    \end{aligned}
\end{equation}
where $\sigma$ is the kernel bandwidth as in Gaussian kernels. 

\item \textbf{Sinh-Gaussian kernels} 
are generalized from Gaussian kernels with an asymmetric skew part $\sinh(x)=(e^x-e^{-x})/2$. The resulted kernel function and its Fourier transform are given as follows,
\begin{equation}
\begin{aligned}\label{k2}
    k_2(\BB{\Delta}) & = \exp(-\frac{\|\BB{\Delta}\|^2}{2\sigma^2})\cdot(1+\sinh{(\BB{\beta}^\top \BB{\Delta})})\\
    \mu_2(\BB{\omega}) &= \big(\frac{\sigma}{\sqrt{2\pi}}\big)^d\exp(-\frac{\sigma^2\|\BB{\omega}\|^2}{2})\big[1-\mathsf{i}\cdot\exp(\frac{\sigma^2\|\BB{\beta}\|^2}{2})\sin(\sigma^2\BB{\beta}^\top\BB{\omega}) \big]\,.
\end{aligned}
\end{equation}

\item \textbf{Cosh-Gaussian kernels} 
are generalized from Gaussian kernels with an asymmetric part, i.e., $\exp(x)$. The resulted kernel function and its Fourier transform are given as follows, 
\begin{equation}
\begin{aligned}\label{k3}
    k_3(\BB{\Delta}) & = \exp(-\frac{\|\BB{\Delta}\|^2}{2\sigma^2})\cdot\exp(\BB{\beta}^\top \BB{\Delta})\\
    \mu_3(\BB{\omega}) &= \big(\frac{\sigma}{\sqrt{2\pi}}\big)^d\exp(-\frac{\sigma^2\|\BB{\omega}\|^2}{2})\exp(\frac{\sigma^2\|\BB{\beta}\|^2}{2})\exp(-\mathsf{i}\cdot\sigma^2\BB{\beta}^\top\BB{\omega})\,.
\end{aligned}
\end{equation}
\end{itemize}

With the specific kernels, we 
can better illustrate the process of estimating 
AsK-RFFs based on the total masses in \eqref{kernel_approx},
which can be computed in advance as they only depend on the kernel function type.
Notice that the four positive measures forms a complex measure, see (\ref{positive_measures}), defined by the non-smooth operator.
That means, no analytical integral is allowed and thus we in turn focus on numerical integration for the estimation of total masses.

Conducting numerical integration for total masses estimation involves computationally extensive in high dimensions (i.e., $d$ is large). To accelerate the estimation, we approximate the total masses on a subset of training data $\mathcal{X}_s$ and formulate this estimation task as a least squares regression problem under the constraints (\ref{constrain_complex_measures_algo}). 
To be specific: denote the number of sub-training samples as $N_s=\vert\mathcal{X}_s\vert$ and $\BB{K}\in\Real^{N_s\times N_s}$ be a sub-Gram matrix. Let $\BB{\omega}$,$\BB{\zeta}$ and $\BB{\nu}$ be as in (\ref{kernel_approx}), denote $\xi_1:=\|\mu_R^+\|$, $\xi_2:=\|\mu_R^-\|$ and $\xi_3:=\|\mu_I^+\|$. We focus on the following optimization problem
\begin{equation}\label{optimization_totalmasses}
    \begin{split}
    &\min_{\xi_1,\xi_2,\xi_3} \,\, \|\BB{K}-\xi_1\cdot\phi\left(\BB{\omega},\mathcal{X}_s\right)\phi\left(\BB{\omega},\mathcal{X}_s\right)^\top +\xi_2\cdot \phi\left(\BB{\zeta},\mathcal{X}_s\right)\phi\left(\BB{\zeta},\mathcal{X}_s\right)^\top \\
    &\qquad\qquad-2\xi_3\cdot\phi\left(\BB{\nu},\mathcal{X}_s\right)\psi\left(\BB{\nu},\mathcal{X}_s\right)^\top\|_F^2\\
    &\rm{s.t.}\quad \begin{array}{lc}
    \xi_1-\xi_2 = k(\BB{0}),~  \xi_1, \xi_2, \xi_3 \geq 0,\,\\
    \end{array}
    \end{split}
\end{equation}
where 
\begin{equation*}
    \left\{
    \begin{aligned}{}
         &\phi(\BB{\omega},\mathcal{X}_s):=\left[\phi(\BB{\omega},\BB{x_1}),\cdots,\phi(\BB{\omega},\BB{x_{N_s}})\right]^{\!\top}  \\
         &\psi(\BB{\nu},\mathcal{X}_s):=\left[\psi(\BB{\nu},\BB{x_1}),\cdots,\psi(\BB{\nu},\BB{x_{N_s}})\right]^{\!\top}\,.\\
    \end{aligned}
    \right.
\end{equation*}
The optimization problem converts the original $d$ dimensional integral problem to a least square problem with $N_s$ samples, and thus the computational complexity is $\mathcal{O}(N_s^3)$, determined by $N_s$. 
We remark that, the total masses do not need to be known before sampling $\BB{\omega}, \BB{\zeta}, \BB{\nu}$, since a modified acceptance-rejection algorithm that samples random variables (r.v.) $\BB{\omega}$ from a finite positive measure $f/\|f\|$ works as follows:
\begin{enumerate}
    \item Given a proposal distribution $g/\|g\|$ satisfying $\sup_{\BB{\omega}}\frac{f(\BB{\omega})}{g(\BB{\omega})}\leq c$ where $c$ is a constant.
    \item Sample a r.v. $\BB{\zeta}$ distributed as $g/\|g\|$.
    \item Sample a r.v. $U$ distributed as a uniform distribution $\mathcal{U}[0,1]$.
    \item If $U\leq\frac{f(\BB{\zeta})}{cg(\BB{\zeta})}$ then set $\BB{\omega}=\BB{\zeta}$; otherwise go back to the step 2.
\end{enumerate}
The proof that the acceptance-rejection algorithm works is deferred to Appendix \ref{append:accpetance_rejection}. The total mass of 
$g$ can be calculated analytically if $g$ is a good measure such as the Gaussian distribution. As shown in the above steps, $\|f\|$ is not required to be necessarily known before the acceptance-rejection sampling. Thus, we can sample the frequency components before estimating total masses. The AsK-RFFs method with the estimation total masses process is shown in Algorithm \ref{alg:B}.
\begin{algorithm}[htb]
\renewcommand{\algorithmicrequire}{\textbf{Input:}}
\renewcommand{\algorithmicensure}{\textbf{Output:}}
\caption{AsK-RFFs with total masses estimation.}
\begin{algorithmic}[1]
\Require A kernel function $k(\BB{x}-\BB{y})$, \#random features $M$ and a sub-training set $\mathcal{X}_s$.
\Ensure The random Fourier features $\phi,\psi:\Real^d\times\Real^d\rightarrow\Real^{2M}$.
\State Obtain a complex measure $\mu=\mu_R+\mathsf{i}\mu_I$ of the kernel $k$ via the Fourier transform.
\State Let $\mu_R=\mu_R^+-\mu_R^-$ and $\mu_I=\mu_I^+-\mu_I^-$ be the Jordan decomposition with four finite positive measures $\mu_R^+$, $\mu_R^-$, $\mu_I^+$ and $\mu_I^-$. Sample $\{\omega_k\}_{k=1}^M\sim \mu_R^+$, $\{\zeta_k\}_{k=1}^M\sim \mu_R^-$ and $\{\nu_k\}_{k=1}^M\sim \mu_I^+$.
\State Randomly sample $N_s=\vert\mathcal{X}_s\vert$ data points and optimize the total masses $\|\mu_R^+\|$, $\|\mu_R^-\|$, $\|\mu_I^+\|$ according to (\ref{optimization_totalmasses}).
\State Return explicit feature mappings $\phi,\psi$ given in (\ref{random_features}).
\end{algorithmic} \label{alg:B}  
\end{algorithm}

\section{Experiments}\label{sec:experiment}
This section will numerically evaluate the proposed AsK-RFFs 
from both kernel approximation and classification performance. Four benchmark datasets are involved, including spambase\footnote{https://archive.ics.uci.edu/ml/datasets.php}\citep{Dua_2019}, ijcnn1\footnote{https://www.csie.ntu.edu.tw/~cjlin/libsvmtools/datasets/\label{libsvm}}, letter\footref{libsvm}, and cod-RNA\footref{libsvm}\citep{chang2011libsvm}. In pre-processing, all data are normalized to $[0,1]^d$ 
and have been randomly partitioned into training and testing set; see 
Table \ref{Asy_vs_sy} for the statistics.
We will evaluate the kernel approximation and classification performance of AsK-RFF for three asymmetric kernels 
(\ref{k1}-\ref{k3}) 
with hyperparameters $\BB{r}=2/d\times\BB{1}$, $\sigma=2$, $\BB{\beta} =0.5\pi/d\times\BB{1}$, and \#sub-training $N_s=50$ for Algorithm \ref{alg:B} where $\BB{1}$ is an all 1 vector with dimension $d$. All the reported results are obtained by running over $10$ trials. The experiments are implemented in MATLAB on a PC with Intel i7-10700K CPU (3.8GHz) and 32GB memory.

\subsection{Total masses estimation}
\begin{figure}[ht]
    \centering
    \subfigure{
    \begin{minipage}[t]{0.305\linewidth}
    \centering
    \includegraphics[width=1\linewidth]{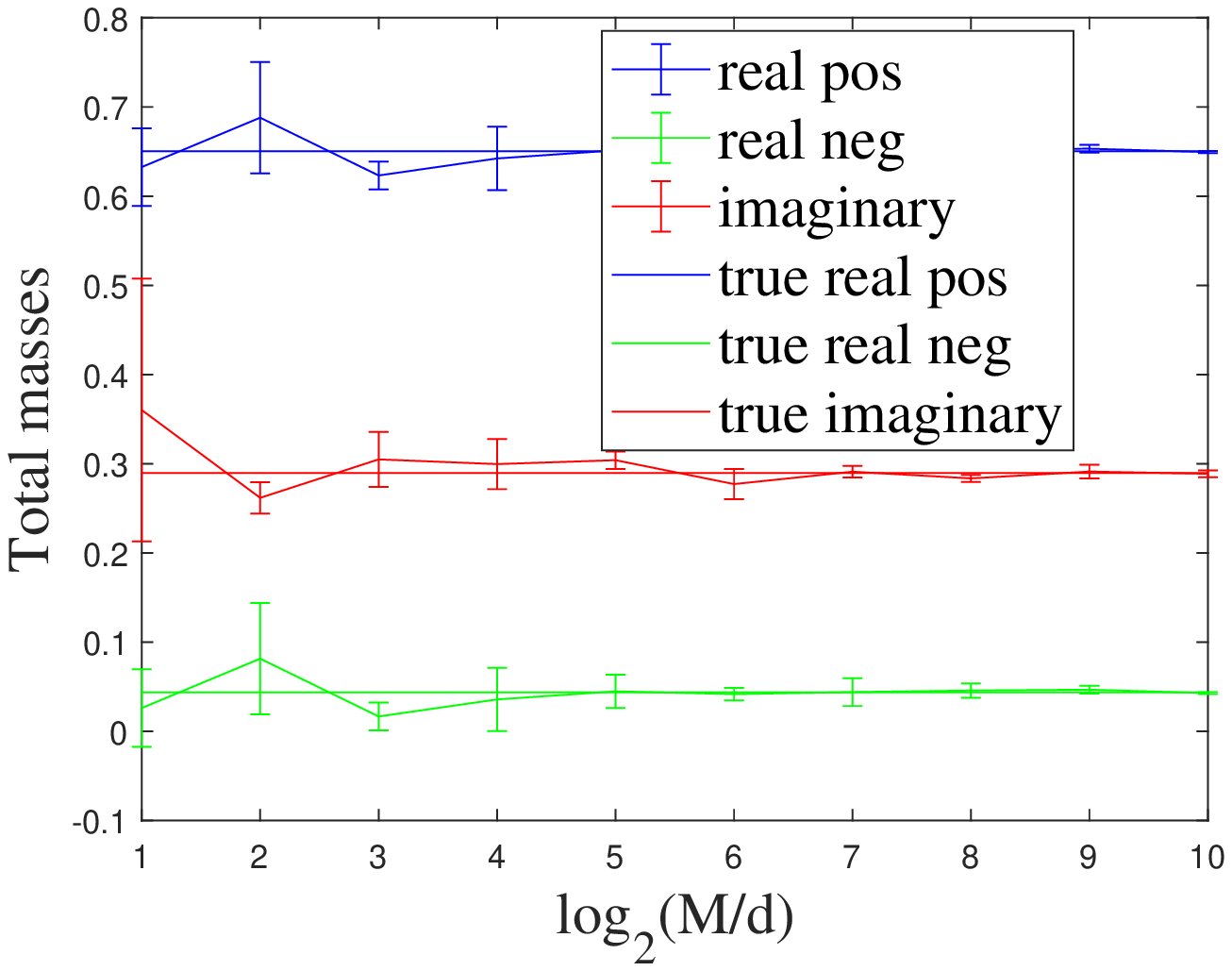}
    \end{minipage}
    }
    \subfigure{
    \begin{minipage}[t]{0.305\linewidth}
    \centering
    \includegraphics[width=1\linewidth]{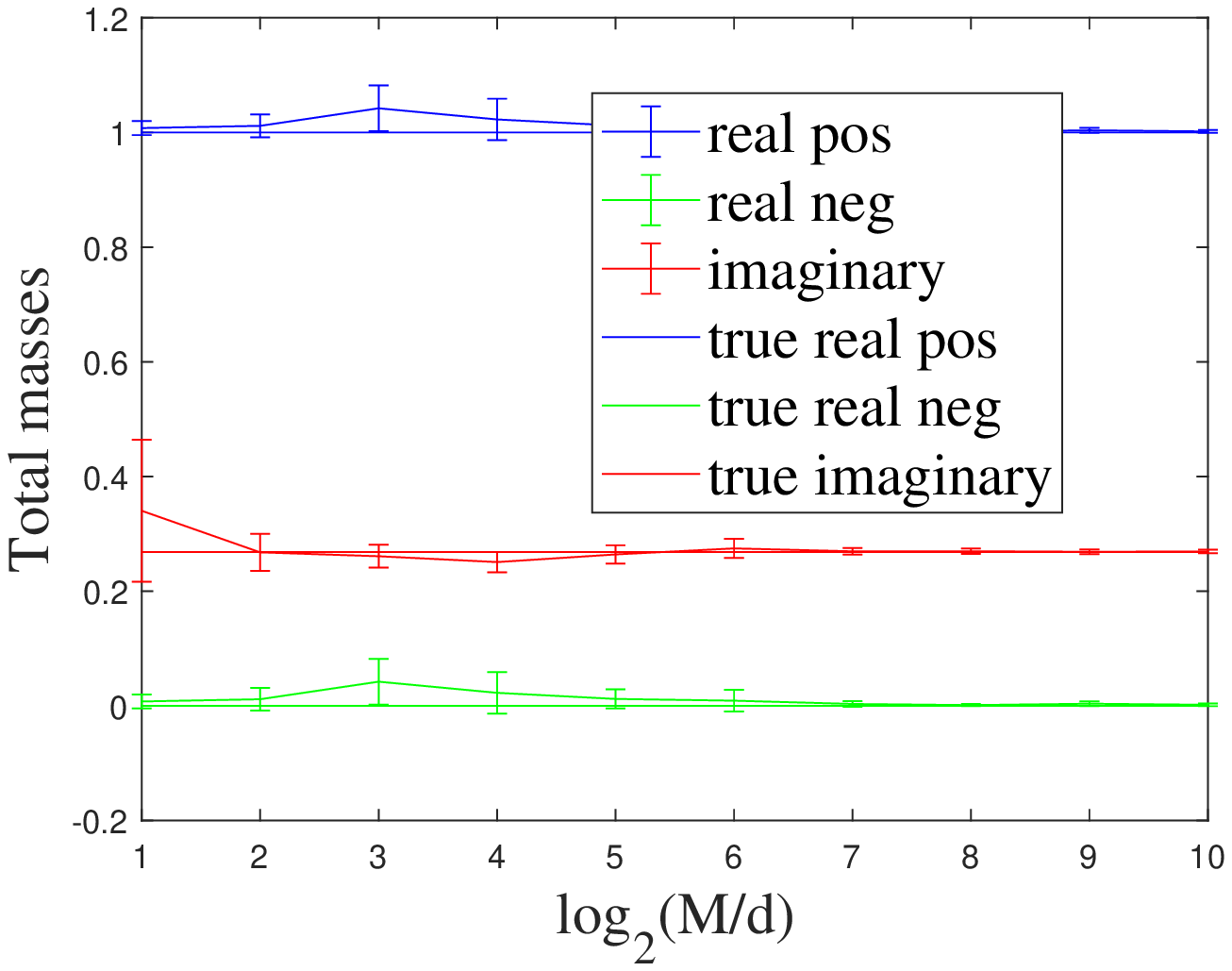}
    \end{minipage}
    }
    \subfigure{
    \begin{minipage}[t]{0.305\linewidth}
    \centering
    \includegraphics[width=1\linewidth]{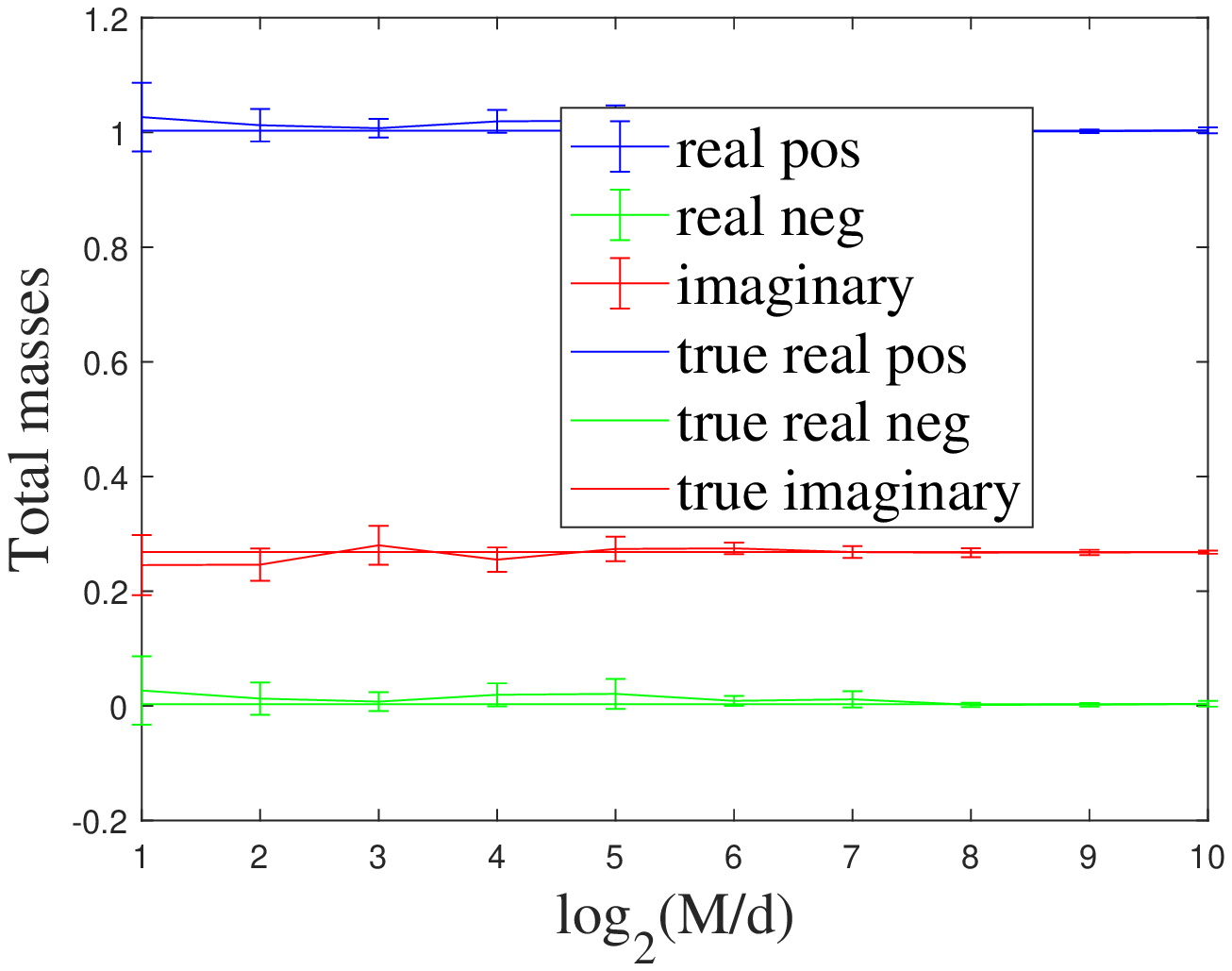}
    \end{minipage}
    }
    \caption{Estimated and accurate total masses of three asymmetric kernels (left: Shift-Gaussian; middle: Sinh-Gaussian; right: Cosh-Gaussian) over 1-dimensional data with hyperparameter $N_s=10$. ``real pos'', ``real neg'' and ``imaginary'' on the legend are the estimation bars with the average and the standard deviation, respectively. ``true real pos'', ``true real neg'' and ``true imaginary'' on the legend are the accurate total masses, respectively. We report the average and the standard deviation of the estimation result (\ref{optimization_totalmasses}) over 10 trials, against \#random features $\log_2(M/d)$.}
    \label{fig:estimated_tv}
\end{figure}

We first estimate 
the total masses with $N_s=10$ on $2000$ random points sampled from a $1$ dimensional Gaussian distribution as a toy example. In this case, the total mass can be accurately calculated and then can be used to evaluate the performance of our estimation.  
The results is plotted in Figure \ref{fig:estimated_tv}, where the solid lines are the accurate total masses calculated by numerical integral. The bars show the average and standard deviation of estimated total masses. 
Overall, the estimated total masses stabilize around the 
accurate total masses and the standard deviation decrease as \#random features increases. The proposed optimization provides a valid estimation for total masses even if $N_s$ is sufficiently small, which illustrates the effectiveness of our estimation method.

\subsection{Kernel approximation}
\begin{figure}[ht]
    \centering
    \subfigure[spambase]{
    \begin{minipage}[t]{0.22\linewidth}
    \centering
    \includegraphics[width=1\linewidth]{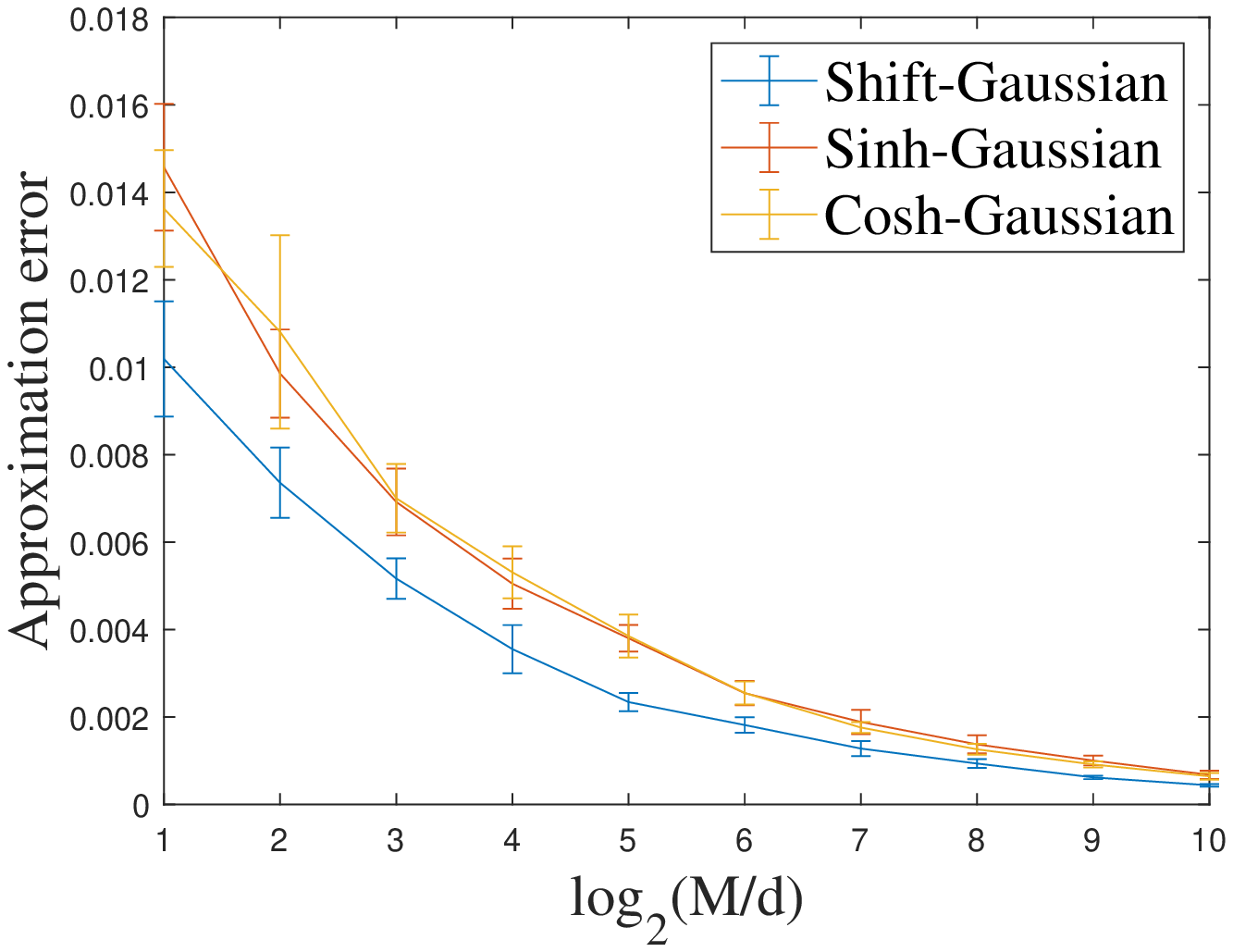}
    \end{minipage}
    }
    \subfigure[letter]{
    \begin{minipage}[t]{0.22\linewidth}
    \centering
    \includegraphics[width=1\linewidth]{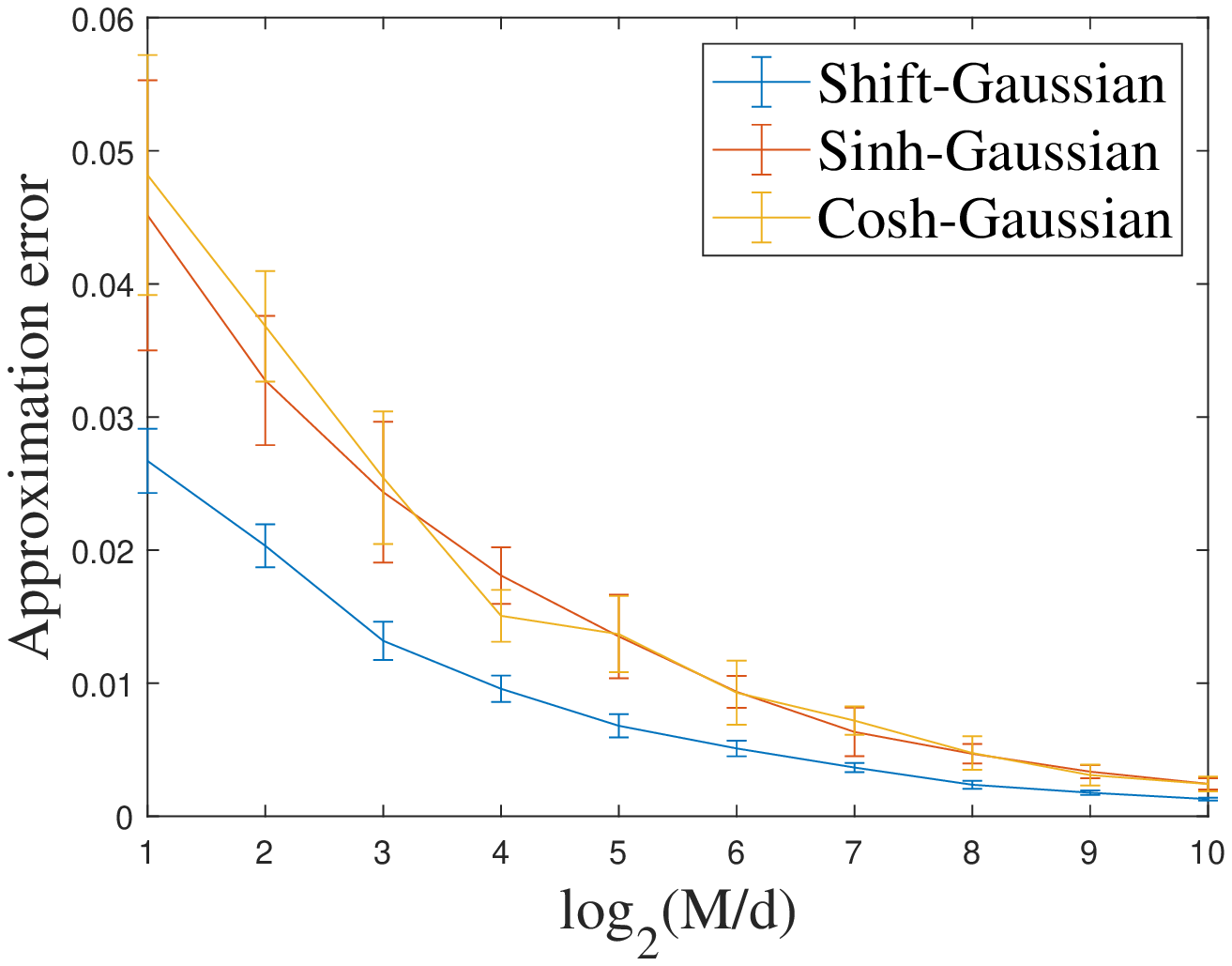}
    \end{minipage}
    }
    \subfigure[ijcnn1]{
    \begin{minipage}[t]{0.22\linewidth}
    \centering
    \includegraphics[width=1\linewidth]{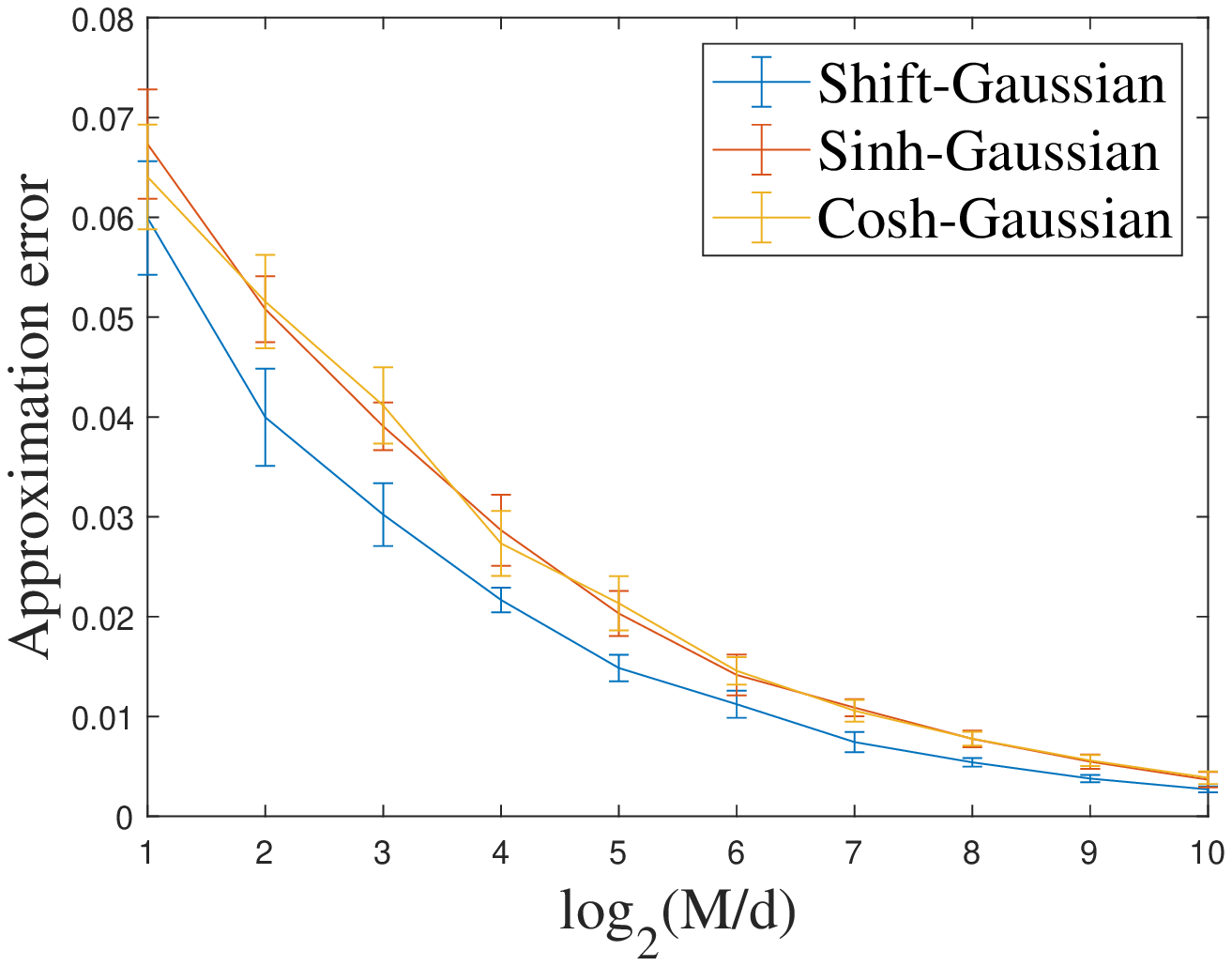}
    \end{minipage}
    }
    \subfigure[cod-RNA]{
    \begin{minipage}[t]{0.22\linewidth}
    \centering
    \includegraphics[width=1\linewidth]{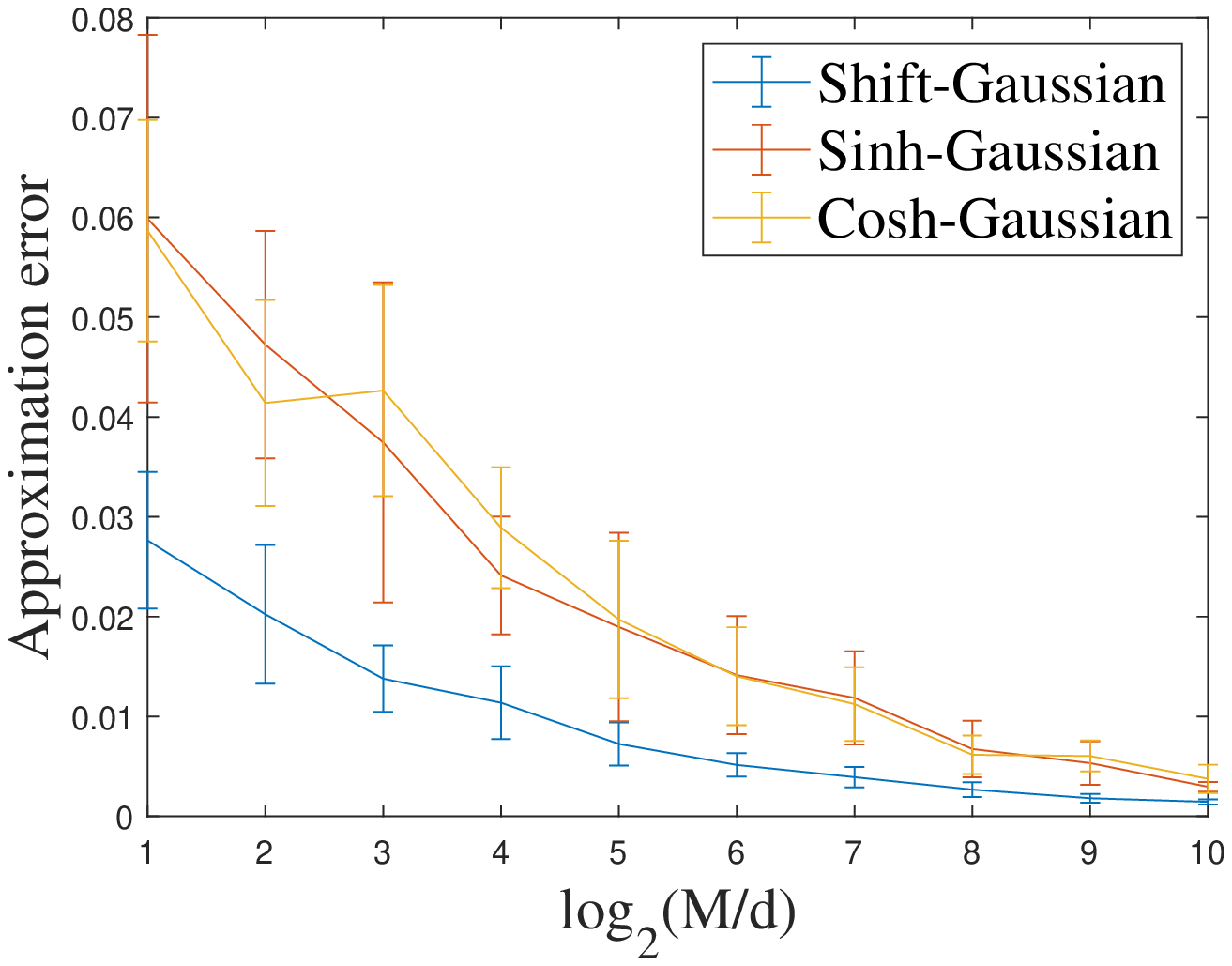}
    \end{minipage}
    }
    \caption{The approximation error bars across three asymmetric kernels on the four benchmark datasets. For each dataset, we run Algorithm \ref{alg:B} with $N_s=50$ over 10 trials and plot the average and the standard deviation of the approximation error against \#random features $\log_2(M/d)$.}
    \label{fig:approximation}
\end{figure}

Next, we evaluate the kernel approximation performance of AsK-RFFs. 
Suppose $\BB{K}$ and $\widetilde{\bm K}$ are the exact Gram matrix and the approximated Gram matrix on $1000$ random selected samples, respectively. 
We use the relative error $\|\BB{K}-\widetilde{\bm K}\|_F/\|\BB{K}\|_F$ to measure the approximation error, where $\|\cdot\|_F$ denotes the Frobenius norm. Figure \ref{fig:approximation} demonstrate 
the approximation error bars for three asymmetric kernels in terms of \#random features $M=d\times[2^{1},2^{2},\cdots,2^9, 2^{10}]$ on four datasets. It is clear that the average approximation error of AsK-RFFs decreases as $M$ increases on the four datasets and 
AsK-RFFs can effectively approximate asymmetric kernel functions. Furthermore, as $M$ increases, the 
approximation performance becomes more stable, i.e., 
the standard deviations reduce significantly. The experimental results perfectly coincide  with 
Proposition \ref{Proposition1}.

\subsection{Classification}
After obtaining the explicit feature mapping, one can apply linear classifiers, e.g.,  Liblinear \citep{fan2008liblinear}, for nonlinear tasks. 
According to the AsK-RFFs method shown in (\ref{kernel_approx}) and (\ref{random_features}), we can obtain a pair of features for each asymmetric kernel function.
In this paper, we concatenate two random features as an explicit feature mapping 
$
    \Phi(x)=\left[\sqrt{\|\mu_{R}^+\|}\phi(\BB{\omega},\BB{x})^\top, \mathsf{i}\sqrt{\|\mu_{R}^-\|}\phi(\BB{\zeta},\BB{x})^\top, \sqrt{2\|\mu_{I}^+\|}\phi(\BB{\nu},\BB{x})^\top,\sqrt{2\|\mu_I^+\|}\psi(\BB{\nu},\BB{x})^\top\right]^\top
$
which is then send to 
LibLinear. 
The hyperparameter in Liblinear is tuned by 5-folds cross-validation on a parameter grid $C=\left[2^{-5},2^{-4},\cdots,2^4, 2^5\right]$. 

\begin{figure}[ht]
    \centering
    \subfigure{
    \begin{minipage}[t]{0.31\linewidth}
    \centering
    \includegraphics[width=1\linewidth]{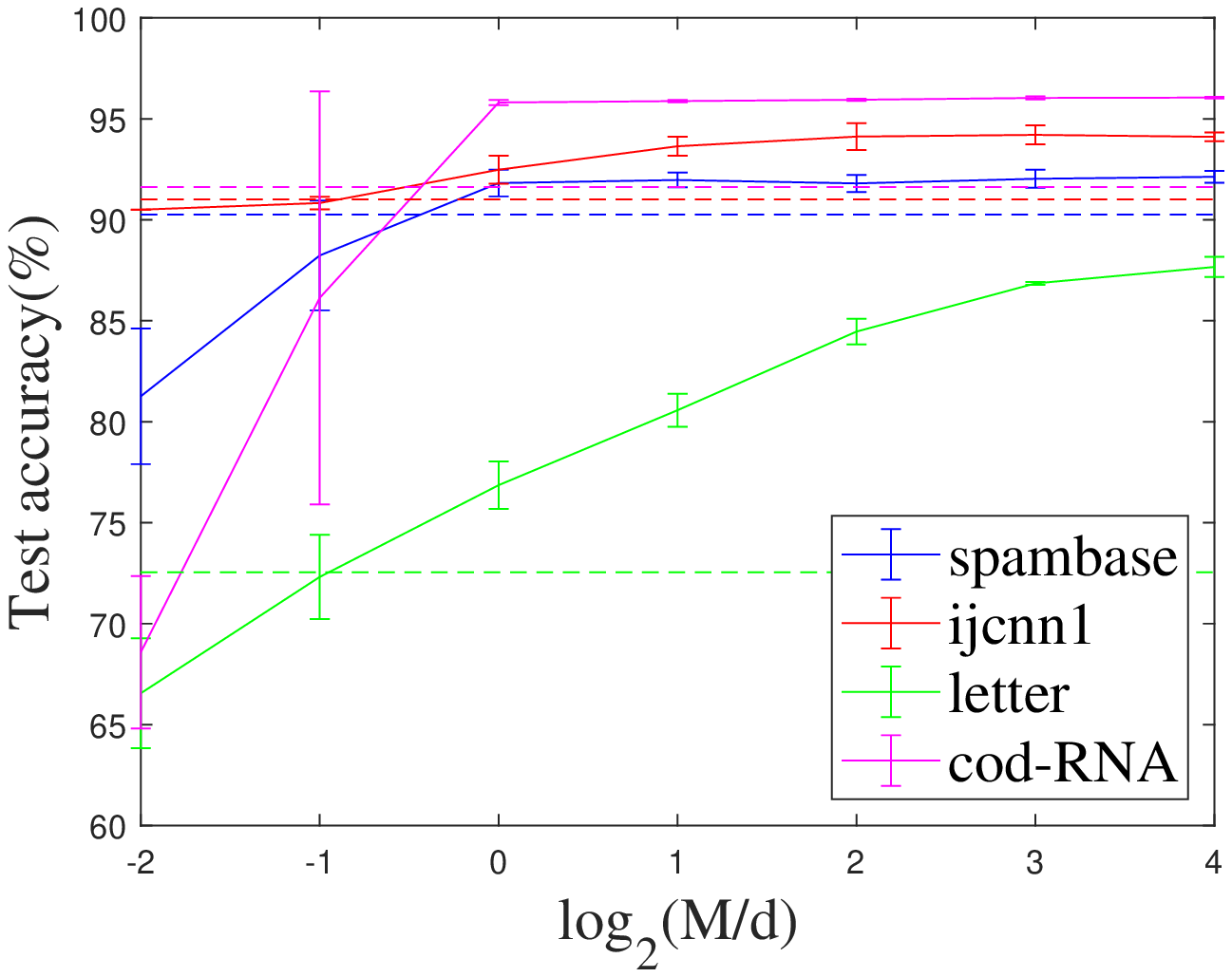}
    \end{minipage}
    }
    \subfigure{
    \begin{minipage}[t]{0.305\linewidth}
    \centering
    \includegraphics[width=1\linewidth]{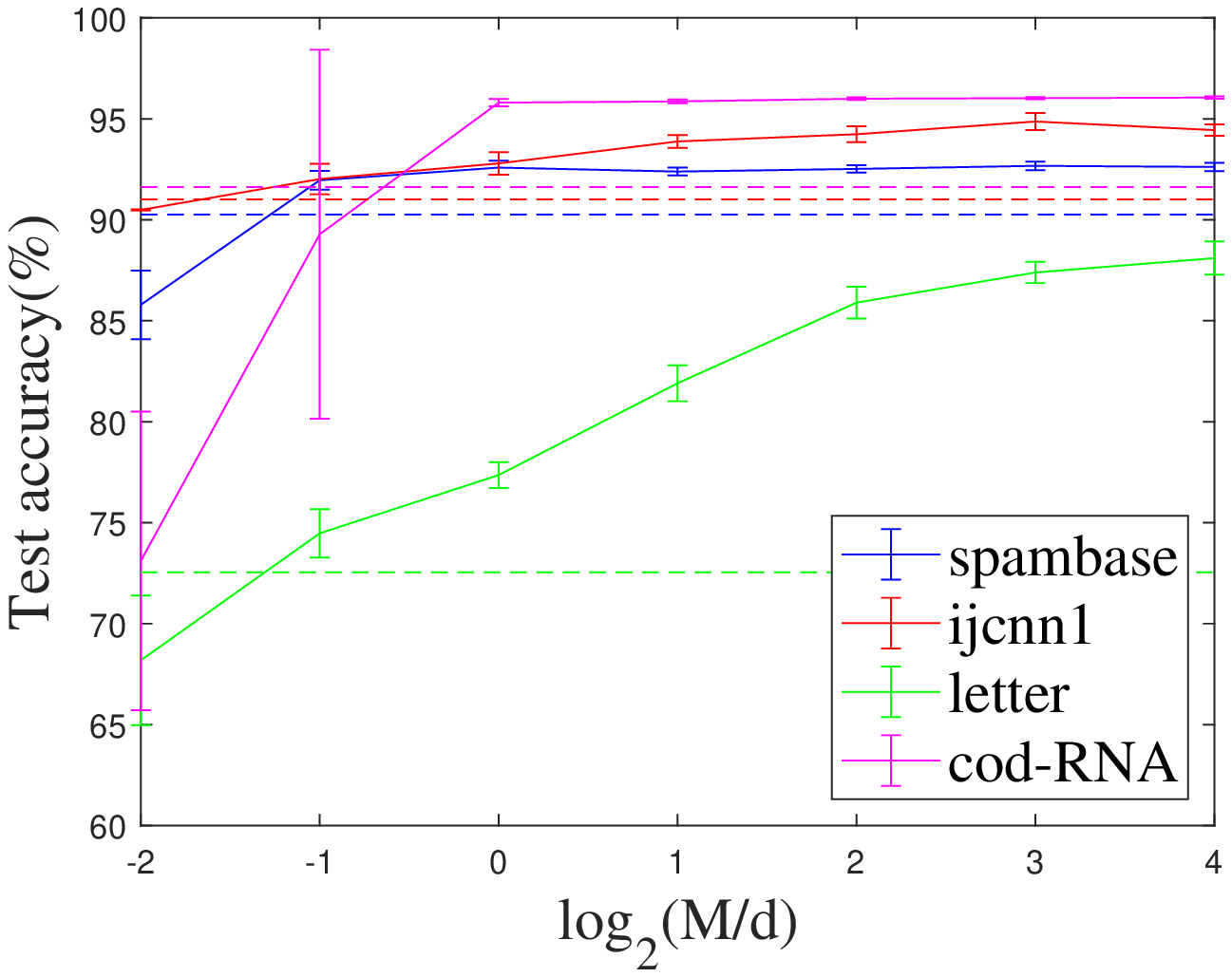}
    \end{minipage}
    }
    \subfigure{
    \begin{minipage}[t]{0.305\linewidth}
    \centering
    \includegraphics[width=1\linewidth]{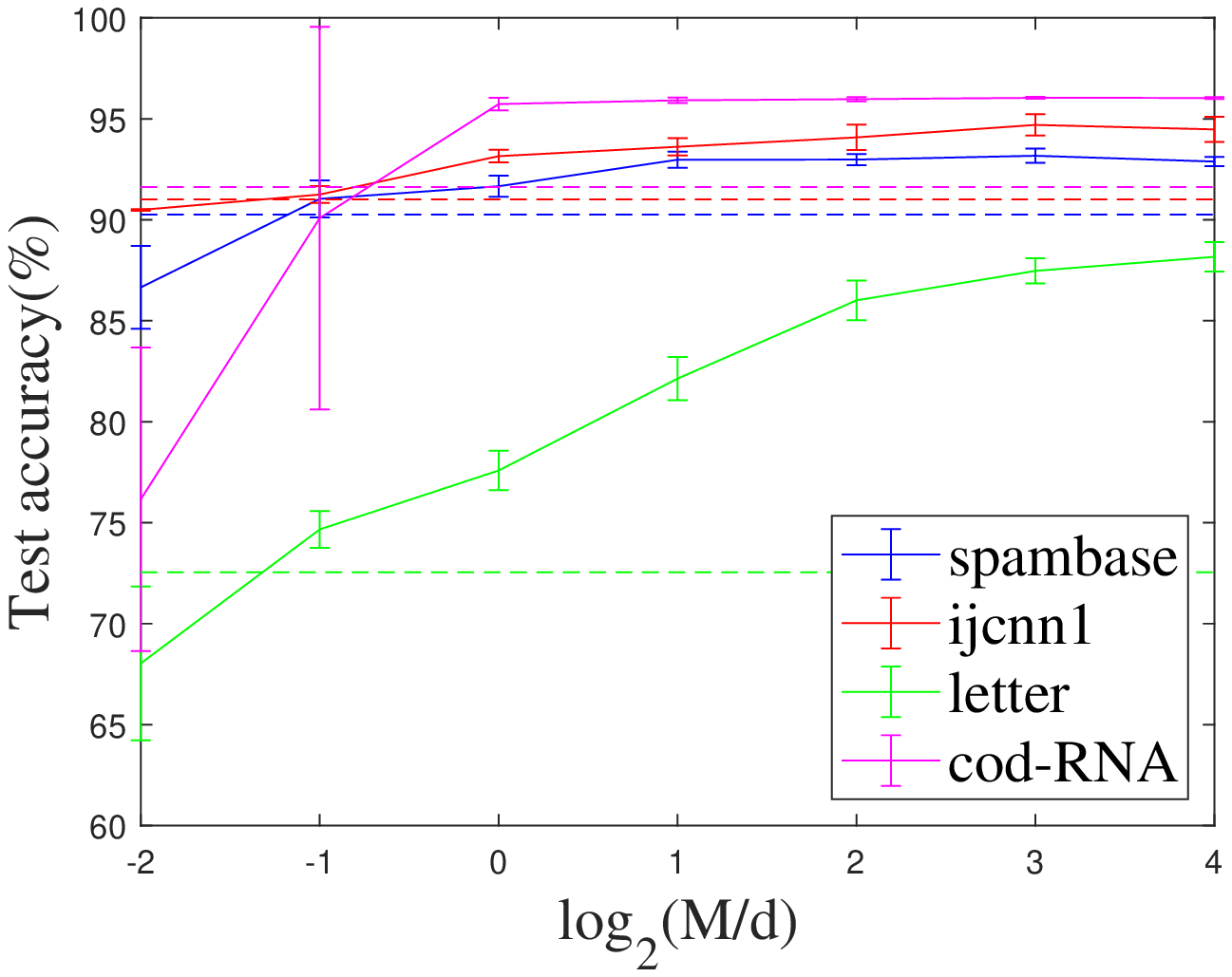}
    \end{minipage}
    }
    
    \setcounter{subfigure}{0}
    \subfigure{
    \begin{minipage}[t]{0.305\linewidth}
    \centering
    \includegraphics[width=1\linewidth]{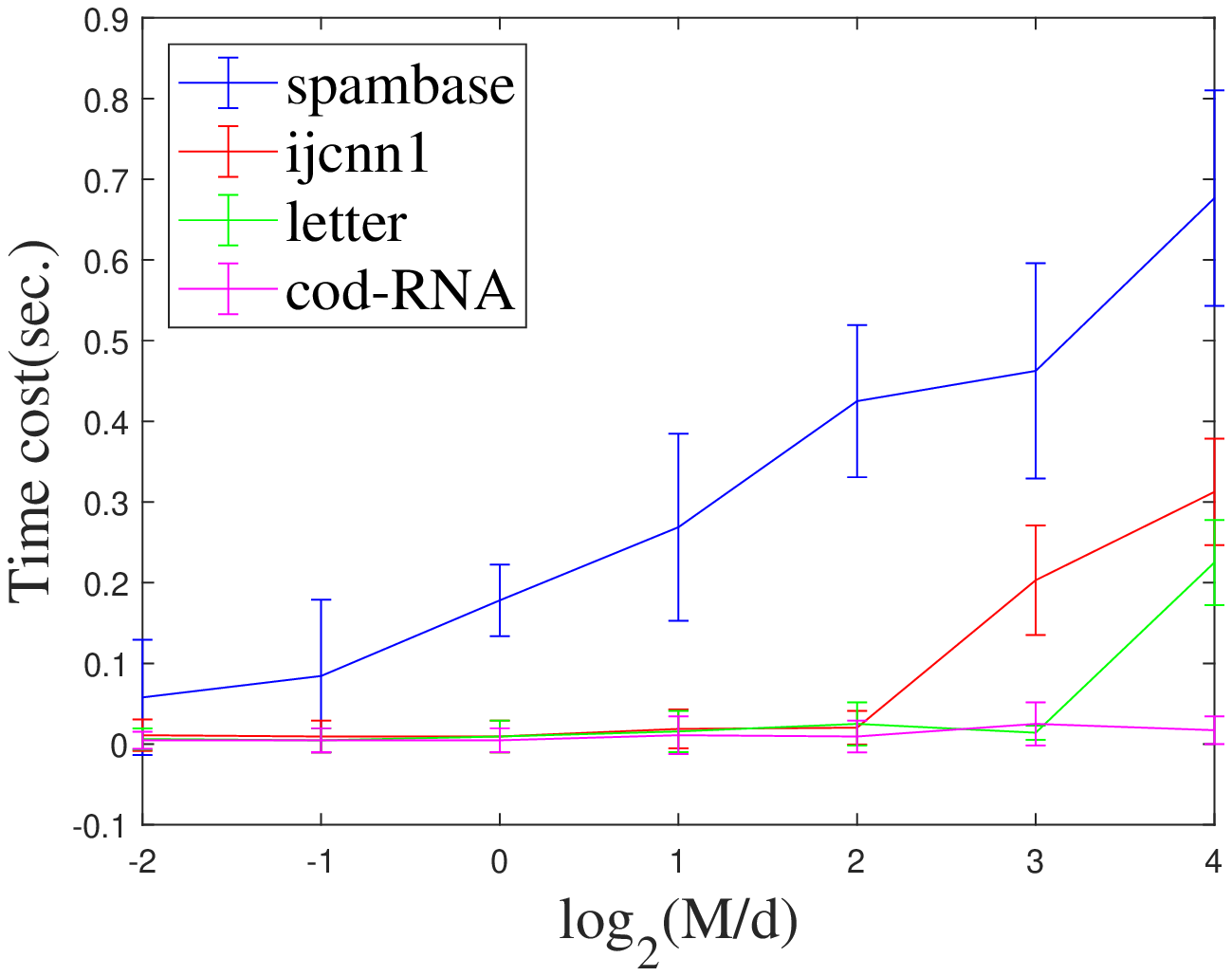}
    \end{minipage}
    }
    \subfigure{
    \begin{minipage}[t]{0.31\linewidth}
    \centering
    \includegraphics[width=1\linewidth]{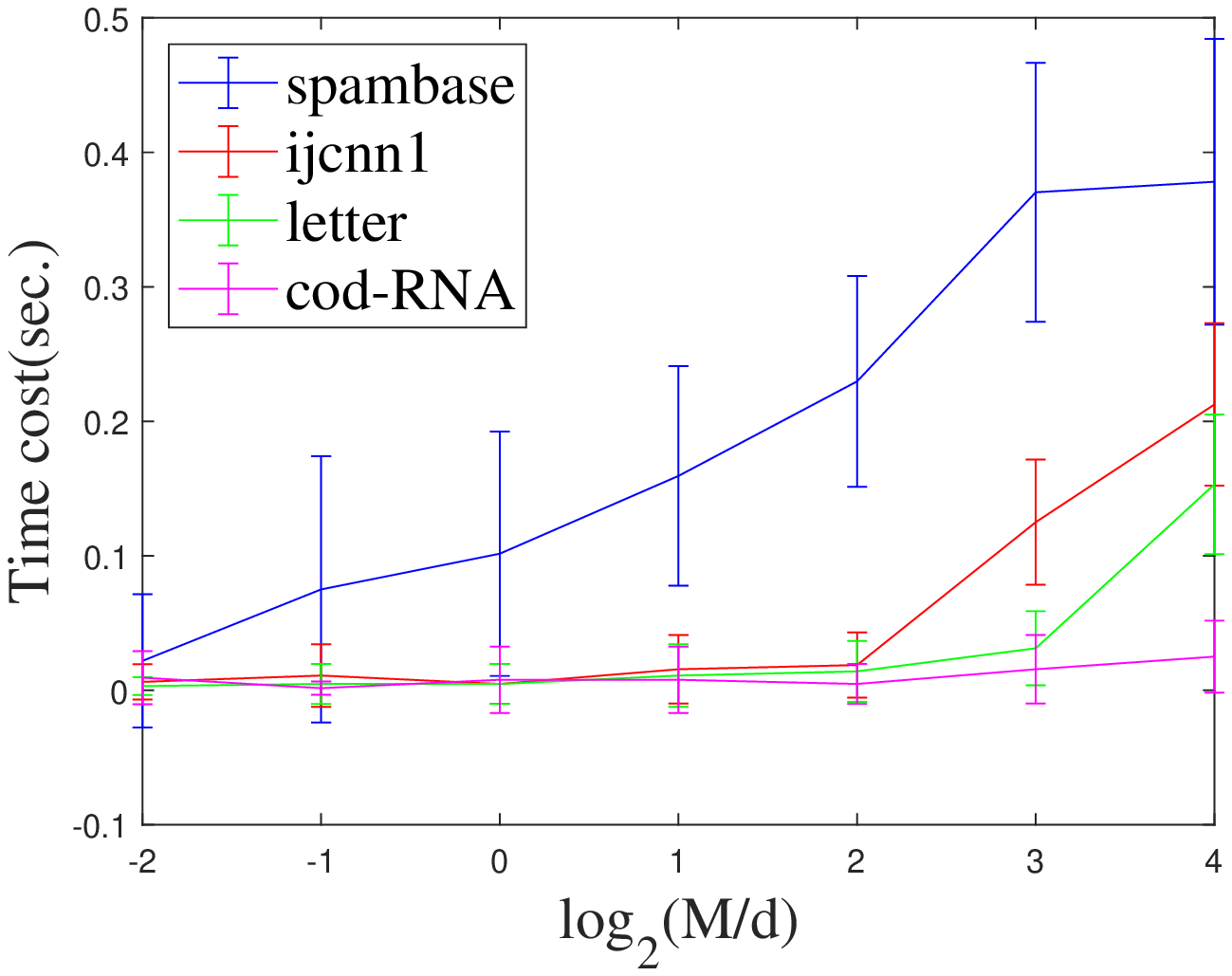}
    \end{minipage}
    }
    \subfigure{
    \begin{minipage}[t]{0.305\linewidth}
    \centering
    \includegraphics[width=1\linewidth]{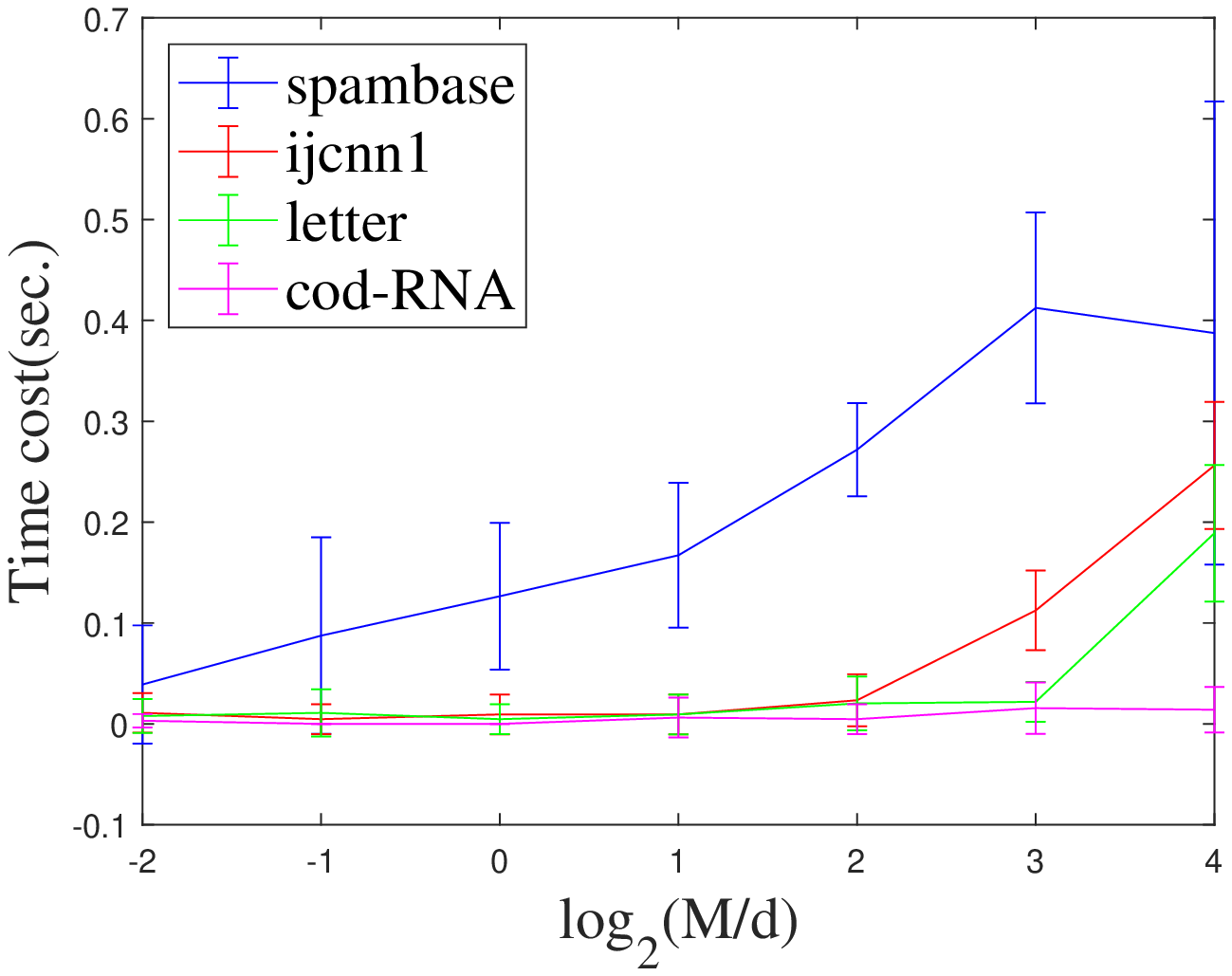}
    \end{minipage}
    }
    \caption{Comparisons of asymmetric kernels for the classification accuracy and sampling time cost with LibLinear on the four datasets. For each asymmetric kernel (left: Shift-Gaussian; middle: Sinh-Gaussian; right: Cosh-Gaussian), we run Algorithm \ref{alg:B} with $N_s=50$ over 10 trials and plot the test accuracy bars and sampling time cost bars against \#random features $\log_2(M/d)$.}
    \label{fig:acc_time}
\end{figure}

Test accuracy and sampling time cost across three asymmetric kernels on the four datasets are illustrated in Figure \ref{fig:acc_time}. 
The accuracy of a linear classifier is given by dashed lines for reference and the solid bars reflect how well the AsK-RFFs extracts information from nonlinear kernels. It also could be found that with more random features, the accuracy is generally higher until saturation and the AsK-RFFs performs better than the linear classifier when $M\geq d$. But with more features, the AsK-RFF needs more computational time, which, however, is always significantly cheaper to calculate the kernel function and solve SVMs in the dual space.


\subsection{Asymmetric kernels vs symmetric kernels}
The above experiments verified that AsK-RFF can efficiently approximate asymmetric kernels in the view of both kernel approximation and classification. Another possible way of handling asymmetric kernels is to make them symmetric, e.g., $\frac{ K^\top + K}{2}$, and then extract regular RFFs from the symmetric part. Here in this part, we compare the performance of AsK-RFF and RFF of the above symmetrization method. Additionally, symmetric kernels, including the linear and the radial basis kernel (RBF), are compared as well. 
Here, the number of random features is 
$M=2d$ and the bandwidth of the RBF kernel is $2$.

The results are reported in Table \ref{Asy_vs_sy} where ``Asy'' and ``Sym'' denote the asymmetric kernel and its symmetric part, respectively. Generally, Ask-RFFs 
output a higher classification accuracy than their symmetric parts, the RBF kernel, and the linear kernel. The better performance over RBF, especially there is nearly $5\%$ improvement on ``letter'', shows that when asymmetric information is crucial, asymmetric kernels may be good choice. The better performance over RFF on the symmetric part, a.k. ``Sym'', verifies the effectiveness of AsK-RFF for approximating asymmetric kernels. 

\begin{table}[ht]
\centering\footnotesize
\setlength{\abovecaptionskip}{0pt}%
\setlength{\belowcaptionskip}{5pt}%
\caption{Test accuracy (the average and the standard deviation over 10 tirals) for different kernels and different approximation methods. 
}\label{Asy_vs_sy}
\begin{tabular}{cccccc}
\specialrule{0.10em}{3pt}{3pt}
\multicolumn{2}{c}{Datasets}   & spambase & ijcnn1 & letter & cod-RNA \\
\specialrule{0.05em}{3pt}{3pt}
\multicolumn{2}{c}{Dimension} &57 &16 &22 &8 \\
\specialrule{0.05em}{3pt}{3pt}
\multicolumn{2}{c}{\#Train} &2760 & 12000 &49990 &59535 \\
\specialrule{0.05em}{3pt}{3pt}
\multicolumn{2}{c}{\#Test} &1841 &6000 &91701 &271617 \\
\specialrule{0.05em}{3pt}{3pt}
Shift-& Asy   &    \textbf{  92.689$\pm$0.207 }   &   \textbf{93.763$\pm$0.599}    &     \textbf{80.631$\pm$1.269}   &    \textbf{95.881$\pm$0.122}     \\
\cmidrule(r){2-6}
Gaussian& Sym   &    92.591$\pm$0.228    &    93.180$\pm$0.614    &    77.975$\pm$1.229    &    95.789$\pm$0.008     \\
\specialrule{0.05em}{3pt}{3pt}
Sinh-& Asy   &     \textbf{92.787$\pm$0.351}     &   \textbf{93.683$\pm$0.481}     &   \textbf{82.455$\pm$0.851}     &    \textbf{95.932$\pm$0.001}     \\
\cmidrule(r){2-6}
Gaussian& Sym   &      92.624$\pm$0.247    &    93.110$\pm$0.587    &    78.553$\pm$0.806    &   95.866$\pm$0.011      \\
\specialrule{0.05em}{3pt}{3pt}                     
Cosh-& Asy   &     \textbf{92.787$\pm$0.210}     &     \textbf{93.583$\pm$0.278}   &    \textbf{82.237$\pm$1.206}    &    \textbf{95.95$\pm$0.038}     \\
\cmidrule(r){2-6}
Gaussian& Sym   &    92.520$\pm$0.210      &    93.464$\pm$0.406    &   77.652$\pm$1.585     &   95.794$\pm$0.008      \\
\specialrule{0.05em}{3pt}{3pt}
\multicolumn{2}{c}{RBF}       &     92.461$\pm$0.300     &    93.157$\pm$0.548    &    77.547$\pm$1.158    &    95.831$\pm$0.069     \\\specialrule{0.05em}{3pt}{3pt}
\multicolumn{2}{c}{Linear} &    90.261$\pm$0.120      &   91.011$\pm$0.048     &   72.541$\pm$0.029     &   91.618$\pm$0.002    \\
\specialrule{0.10em}{3pt}{3pt}
\end{tabular}
\end{table}

\section{Conclusion}\label{sec:conclusion}


In this paper, we investigated a unified kernel approximation framework by RFFs which can uniformly approximate asymmetric kernels. A spectral representation of asymmetric kernels via complex measure was presented by the proposed AsK-RFFs method which could be applied to both asymmetric and symmetric kernels including positive definite and indefinite kernels. 
In this framework, we could design kernels by altering their complex measures from a statistics perspective. Besides, we proposed a subset-based fast estimation method for total masses on a sub-training set to speed up our approximation process. 
The effecietnveness of AsK-RFFs was evaluated in experimental results on several large classification datasets. 
Another intererting avenue for asymmetric kernel approximation is the Nystr{\"o}m-type method \citep{Williams01usingthe,drineas2005nystrom}. The Nystr{\"o}m-type approximation process that uses low-rank matrices to approximate the asymmetric kernel matrix is worthy of further research.

\backmatter







\begin{appendices}
\section{Proof of Proposition \ref{Proposition1}}\label{append:proposition1}
The proof strategy closely follows which of \citep{rahimi2007random, error2015rff}. Suppose $k(\BB{\Delta})$ is twice differentiable. Denote $f(\BB{\Delta}):=s(\BB{\Delta})-k(\BB{\Delta})$.  Let $\mathcal{X_{\BB{\Delta}}}=\{\BB{x}-\BB{y}\vert\BB{x},\BB{y}\in\mathcal{X}\}$ which is compact with diameter $l$. An $\epsilon$-net can be found that covers $\mathcal{X_{\BB{\Delta}}}$ with at most $T=(\frac{4l}{r})^d$ balls of radius $r$ \citep{cucker2002mathematical}. Let $\{\BB{\Delta_k}\}_{k=1}^T$ be centers and $L_f$ be the Lipschitz constant of $f$. There is a fact that $\vert f(\BB{\Delta})\vert<\epsilon$ for all $\BB{\Delta}\in\mathcal{X_{\BB{\Delta}}}$ if $\vert f(\BB{\Delta_k})\vert<\frac{\epsilon}{2}$ and $L_f<\frac{\epsilon}{2r}$ for all $k$.

Since $f$ is differentiable, we denote $L_f=\triangledown f(\BB{\Delta}^\star)$ where $\BB{\Delta}^\star=\arg\max_{\BB{\Delta}\in\mathcal{X_{\BB{\Delta}}}} \|\triangledown f(\BB{\Delta})\|$. We have
\begin{equation*}
    \begin{aligned}
        \mathbb{E}[L_f^2]&=\mathbb{E}\|\triangledown f(\BB{\Delta}^\star)\|^2\\
        &\leq \mathbb{E}\|\triangledown s(\BB{\Delta}^\star)\|^2-\mathbb{E}\|\triangledown k(\BB{\Delta}^\star)\|^2\\
        &\leq \mathbb{E}\|\triangledown s(\BB{\Delta}^\star)\|^2\\
        &=\mathbb{E}\big\Vert\frac{1}{M}\triangledown \sum_{l=1}^M\|\mu_R^+\|\cos(\BB{\omega}_l^\top\BB{\Delta}^\star)-\|\mu_R^-\|\cos(\BB{\zeta}_l^\top\BB{\Delta}^\star)-2\|\mu_I^+\|\sin(\BB{\nu}_l^\top\BB{\Delta}^\star)\big\Vert^2\\
        &=\mathbb{E}\big\Vert\triangledown\big[ \|\mu_R^+\|\cos(\BB{\omega}^\top\BB{\Delta}^\star)-\|\mu_R^-\|\cos(\BB{\zeta}^\top\BB{\Delta}^\star)-2\|\mu_I^+\|\sin(\BB{\nu}^\top\BB{\Delta}^\star)\big]\big\Vert^2\\
        &=\mathbb{E}\left\Vert\|\mu_R^+\|\sin(\BB{\omega}^\top\BB{\Delta}^\star)\BB{\omega}-\|\mu_R^-\|\sin(\BB{\zeta}^\top\BB{\Delta}^\star)\BB{\zeta}+2\|\mu_I^+\|\cos(\BB{\nu}^\top\BB{\Delta}^\star)\BB{\nu}\right\Vert^2\\
        &\leq\mathbb{E}\big[\vert\sin(\BB{\omega}^\top\BB{\Delta}^\star)\vert\cdot\|\mu_R^+\|\|\BB{\omega}\|+\vert\sin(\BB{\zeta}^\top\BB{\Delta}^\star)\vert\cdot\|\mu_R^-\|\|\BB{\zeta}\|\\
        &\quad+2\vert\cos(\BB{\nu}^\top\BB{\Delta}^\star)\vert\cdot\|\mu_I^+\|\|\BB{\nu}\|\big]^2\\
        &\leq\mathbb{E}\big[\|\mu_R^+\|\cdot\|\BB{\omega}\|+\|\mu_R^-\|\cdot\|\BB{\zeta}\|+2\|\mu_I^+\|\cdot\|\BB{\nu}\|\big]^2\\
        &\leq\mathbb{E}\left[\left(\|\mu_R^+\|^2+\|\mu_R^-\|^2+(\sqrt{2}\|\mu_I^+\|)^2 \right)\cdot\left( \|\BB{\omega}\|^2+\|\BB{\zeta}\|^2+(\sqrt{2}\|\BB{\nu}\|)^2\right) \right]\\
        &=\left(\|\mu_R^+\|^2+\|\mu_R^-\|^2+2\|\mu_I^+\|^2 \right)\cdot\mathbb{E}\left( \|\BB{\omega}\|^2+\|\BB{\zeta}\|^2+2\|\BB{\nu}\|^2\right):=\alpha_\mu^2\,.
    \end{aligned}
\end{equation*}
The first and second inequalities are derived in a similar fashion with \citep{error2015rff}. The third inequality is from the triangle inequality  $\|a+b+c\|\leq\|a\|+\|b\|+\|c\|$ with $a,b,c$ being vectors. The fourth inequality holds because the sine and cosine functions are bounded by $\pm 1$. The fifth inequality is from the Cauchy–Schwarz inequality $(\sum_{i=1}^n u_i v_i)^2 \leq (\sum_{i=1}^n u_i^2)(\sum_{i=1}^n v_i^2)$ with $u_i, v_i$ being real numbers.

According to the Markov’s inequality, we obtain a probability bound for the Lipschitz constant,
\begin{equation}\label{bound1}
{\rm Pr}\left(L_f\geq\frac{\epsilon}{2r}\right)={\rm Pr}\left(L_f^2\geq(\frac{\epsilon}{2r})^2\right)\leq\alpha_\mu^2\left(\frac{2r}{\epsilon}\right)^2\,.
\end{equation}
Given a fixed $\BB{\Delta}$, $s(\BB{\Delta})$ is bounded by $\pm\left(\|\mu_R^+\|+\|\mu_R^-\|+2\|\mu_I^+\|\right)$ with the expectation $k(\BB{\Delta})$. Applying the Hoeffding’s
inequality and a union bound to anchors in the $\epsilon$-net, we have
\begin{equation}\label{bound2}
{\rm Pr}\left(\cup_{k=1}^T\vert f(\BB{\Delta_k})\vert\geq\frac{\epsilon}{2}\right)\leq T{\rm Pr}\left(\vert f(\BB{\Delta_k}\right)\vert\geq\frac{\epsilon}{2})\leq 2T\exp\left(-\frac{M\epsilon^2}{8\|\mu\|^2}\right)\,.
\end{equation}
Combining bounds (\ref{bound1}) and (\ref{bound2}), we obtain a bound in terms of $r$,
\begin{equation*}
{\rm Pr}\left(\sup_{\BB{\Delta}\in\mathcal{X_{\BB{\Delta}}}}\vert f(\BB{\Delta})\vert\leq\epsilon\right)\geq 1-\kappa_1 r^{-d}-\kappa_2 r^2,
\end{equation*}
where $\kappa_1 = 2(4l)^d\exp\left(-\frac{M\epsilon^2}{8\|\mu\|^2}\right)$\,, $\kappa_2=4\alpha_\mu^2\epsilon^{-2}$. We choose $r$ such that $d\kappa_1r^{-d-1}-2\kappa_2r=0$ i.e., $r=\left(\frac{d\kappa_1}{2\kappa_2}\right)^{\frac{1}{d+2}}$, as did in \citep{error2015rff}. Then we have a bound
\begin{equation*}
    \begin{aligned}{}
        {\rm Pr}\left(\sup_{\BB{\Delta}\in\mathcal{X_{\BB{\Delta}}}}\vert f(\BB{\Delta})\vert>\epsilon\right)&\leq \left((\frac{d}{2})^\frac{-d}{d+2}+(\frac{d}{2})^\frac{2}{d+2}\right)\cdot2^\frac{6d+2}{d+2}\left(\frac{\alpha_\mu l}{\epsilon}\right)^\frac{2d}{d+2}\exp\left(\frac{-M\epsilon^2}{4(d+2)\|\mu\|^2}\right)\\
        &=\beta_d\left(\frac{\alpha_\mu l}{\epsilon}\right)^\frac{2d}{d+2} \exp\left(\frac{-\epsilon^2 M}{4(d+2)\|\mu\|^2}\right)\,.\\
    \end{aligned}
\end{equation*}
Thus, we can control the pointwise error no more than $\epsilon$ with probability at least $1-\delta$ as long as $M \geq 4(d+2)\|\mu\|^2\epsilon^{-2}\left(\log\frac{\beta_d}{\delta}+\frac{2d}{d+2}\log\frac{\alpha_\mu l}{\epsilon}\right)$. \cite{error2015rff} provide the maximum value $\beta_{64}=66$ and asymptotic values $\lim_{d\rightarrow\infty}\beta_d=64$ and the exponent of the second factor can be loosened to 2 if $\alpha_\mu l\geq\epsilon$.

\section{Proof of the Acceptance-Rejection sampling algorithm works}\label{append:accpetance_rejection}
Denote $F(\BB{t}):=f(\BB{\zeta}\leq\BB{t})$ and $G(\BB{t}):=g(\BB{\zeta}\leq\BB{t})$. Denote two events $A=\{\BB{\zeta}\leq\BB{t}\}$ and $B=\{U\leq\frac{f(\BB{\zeta})}{cg(\BB{\zeta})}\}$. Then, $F/\|f\|$ and $G/\|g\|$ are two probability measures. Suppose a r.v. $\BB{\zeta}$ is sampled from a distribution $G/\|g\|$ and a r.v. $U$ is sampled from a uniform distribution $\mathcal{U}[0,1]$.

We need to prove
${\rm Pr}(A\vert B)=F/\|f\|$. 
To this end, we utilize the Bayes' theorem, i.e., ${\rm Pr}(A\vert B)={\rm Pr}(B\vert A)\cdot{\rm Pr}(A)/{\rm Pr}(B)$. 

Firstly, $\BB{\zeta}$ follows the distribution $G/\|g\|$, then ${\rm Pr}(A)={\rm Pr}(\BB{\zeta}\leq\BB{t})=G(\BB{t})/\|g\|$.
Secondly, the probability ${\rm Pr}(B)$ is given as follows,
\begin{equation*}
    \begin{aligned}
        {\rm Pr}(B)&=\int_{\Real^d} {\rm Pr}\left(U\leq\frac{f(\BB{\zeta})}{cg(\BB{\zeta})}\vert\BB{\zeta}=\BB{t})\right)\cdot{\rm Pr}(\BB{\zeta}=\BB{t})\mathrm{d}\BB{t}\\
        &=\int_{\Real^d} \frac{f(\BB{t})}{cg(\BB{t})}\cdot \frac{g(\BB{t})}{\|g\|}\mathrm{d}\BB{t}\\
        &=\frac{\|f\|}{c\|g\|}.\,
    \end{aligned}
\end{equation*}
Thirdly, the probability ${\rm Pr}(B\vert A)$ can be derived by,
\begin{equation*}
    \begin{aligned}
        {\rm Pr}(B\vert A)&={\rm Pr}\left(U\leq\frac{f(\BB{\zeta})}{cg(\BB{\zeta})}\vert\BB{\zeta}\leq\BB{t}\right)\\
        &=\frac{{\rm Pr}\left(U\leq\frac{f(\BB{\zeta})}{cg(\BB{\zeta})}, \BB{\zeta}\leq\BB{t}\right)}{{\rm Pr}(\BB{\zeta}\leq\BB{t})} \\
        &=\frac{\|g\|}{G(\BB{t})}\int_{[-\infty]^d}^{\BB{t}} {\rm Pr}\left(U\leq\frac{f(\BB{\zeta})}{cg(\BB{\zeta})}\vert\BB{\zeta}=\BB{w}\leq\BB{t}\right)\cdot\frac{g(\BB{w})}{\|g\|}\mathrm{d}\BB{w}\\
        &=\frac{\|g\|}{G(\BB{t})}\int_{[-\infty]^d}^{\BB{t}} \frac{f(\BB{w})}{cg(\BB{w})}\frac{g(\BB{w})}{\|g\|}\mathrm{d}\BB{w}\\
        &=\frac{F(t)}{cG(t)}.\,
    \end{aligned}
\end{equation*}
Then, the probability ${\rm Pr}\left(\BB{\zeta}\leq\BB{t}\vert U\leq\frac{f(\BB{\zeta})}{cg(\BB{\zeta})}\right)={\rm Pr}(A\vert B)$ can be calculated by the following equation.

\begin{equation*}
    \begin{aligned}
        {\rm Pr}\left(\BB{\zeta}\leq\BB{t}\vert U\leq\frac{f(\BB{\zeta})}{cg(\BB{\zeta})}\right)&={\rm Pr}(A\vert B)\\
        &=\frac{{\rm Pr}(B\vert A)\cdot{\rm Pr}(A)}{{\rm Pr}(B)}\\
        &=\frac{F(t)}{cG(t)}\cdot\frac{G(\BB{t})}{\|g\|}\cdot\frac{c\|g\|}{\|f\|}\\
        &=\frac{F(t)}{\|f\|}\,,
    \end{aligned}
\end{equation*}
which concludes the proof.
\end{appendices}

\section*{Declarations}

\begin{itemize}
\item Funding This work was supported by National Natural Science Foundation of China (No. 61977046), Shanghai Municipal Science and Technology Major Project (2021SHZDZX0102).
\item Competing interests The authors declare that they have no known competing financial interests or personal relationships that could have appeared to influence the work
reported in this paper.
\item Ethics approval Not applicable.
\item Consent to participate Not applicable.
\item Consent for publication Not applicable.
\item Availability of data and materials Not applicable.
\item Code availability Not applicable.
\item Authors' contributions Not applicable.
\end{itemize}

\bibliography{sn-bibliography}


\end{document}